%% file: ms.tex
\documentclass[10pt,twocolumn,letterpaper]{article}

\usepackage{3dv}
\usepackage{times}
\usepackage{epsfig}
\usepackage{graphicx}
\usepackage{amsmath}
\usepackage{amssymb}
\usepackage{xspace}
\usepackage{xcolor}   
\usepackage{soul}

\usepackage{shortbold}
\newcommand{\SOT}{{\mathrm{SO}(3)}}
\newcommand{\parnobf}[1]{{\vspace{1mm} \noindent \textbf{{#1}.}}}
\newcommand{\parnoit}[1]{{\vspace{1mm} \noindent \textit{{#1}.}}}
\newcommand{\parnobfnoper}[1]{{\vspace{1mm} \noindent \textbf{{#1}}}}

\usepackage{graphicx}
\usepackage{multirow}
\usepackage{multicol}
\usepackage{array}
\usepackage{verbatim}
\usepackage{booktabs}
\usepackage{mathtools}

\usepackage[pagebackref=true,breaklinks=true,colorlinks,bookmarks=false]{hyperref}

\threedvfinalcopy 

\def\threedvPaperID{64}

\ifthreedvfinal\fi

\makeindex
\begin{document}

\title{The 8-Point Algorithm as an Inductive Bias for Relative Pose Prediction by ViTs}

\author{ Chris Rockwell,~~~Justin Johnson,~~~David F. Fouhey\\University of Michigan}

\maketitle

\begin{abstract}
We present a simple baseline for directly estimating the relative pose
(rotation and translation, including scale) between two images. Deep methods
have recently shown strong progress but often require complex or multi-stage
architectures. We show that a handful of modifications can be applied to a
Vision Transformer (ViT) to bring its computations close to the Eight-Point
Algorithm. This inductive bias enables a simple method to be
competitive in multiple settings, often substantially improving over the
state of the art with strong performance gains in limited data regimes.
\end{abstract}

\section{Introduction}

\input{introduction.tex}

\section{Related Work}

\input{related.tex}

\section{Approach} \label{sec:approach}

\input{approach.tex}

\section{Experiments} \label{sec:experiments}

\input{experiments.tex}

\section{Discussion} \label{sec:discussion}
\input{discussion.tex}

\appendix
\clearpage
\newpage
\section{Appendix}

\noindent Our supplement presents the following.

\parnobf{Additional experimental details for the paper's main results} These consist
of detailed network architectures (\S\ref{sec:arch}), 
descriptions of datasets and data pre-processing (\S\ref{sec:dataset}),
and additional analysis that would not fit in the main paper (\S\ref{sec:analysis}).

\parnobf{Discussion of the Essential Matrix Module} We present additional discussion
and exposition of the Essential Matrix Module. This consists of a more detailed
derivation of the fact that the unique entries of $\UB^\top \UB$ can be written
by $\PhiB^\top \AB \PhiB$ as described in the main paper (\S\ref{sec:derivation}),
how accurately the network can compute $\PhiB^\top \AB \PhiB$ in practice (\S \ref{sec:limitations}), as well as an explicit writing-out of one of the key steps of this (\S\ref{sec:matrixverify}).
It also includes a detailed description of the experiment done with synthetic data
which appears in the method section (\S\ref{sec:synthetic}).

\section{Detailed Network Architectures}
\label{sec:arch}

\input{supp_arch.tex}

\section{Dataset Information}
\label{sec:dataset}
\input{supp_dataset.tex}

\section{Additional Analysis}
\label{sec:analysis}
\input{supp_additional.tex}

\clearpage

\section{Derivation of the Unique Entries of $\UB^\top \UB$}
\label{sec:derivation}
\input{supp_deriv.tex}

\section{Discussion of Limitations}
\label{sec:limitations}

\input{supp_limitations.tex}

\section{Synthetic Experimental Details}
\label{sec:synthetic}
\input{supp_synth.tex}

\onecolumn

\section{Verifying that $\phi(\xB) \phi(\xB')$ contains all the terms needed for $\UB^\top \UB$}
\label{sec:matrixverify}
\input{supp_matrix.tex}

\twocolumn

\clearpage

{\small
\bibliographystyle{ieee_fullname}
\bibliography{local}
}

\end{document}

%% file: introduction.tex
Estimating the relative pose between two images is
a fundamental vision problem~\cite{hartley2003multiple}, with
applications including 3D understanding~\cite{jin2021planar,luo2020consistent,schonberger2016structure}
and extended reality~\cite{lin2021barf,martin2021nerf,mildenhall2020nerf,zhou2018stereo}.
Early work focused on robust~\cite{Fischler81} fitting of
models~\cite{hartley2003multiple,hartley1997defense,longuet1981computer,nister2004efficient}
on detected correspondences ~\cite{bay2006surf,lowe2004distinctive,rublee2011orb} between the images.
This strategy can fail catastrophically with poor correspondence, which is especially frequent
in the {\it wide baseline} setting, when the images have a substantial pose difference.
Moreover, even when it is successful, it cannot recover the {\it scale} of the translation~\cite{hartley2003multiple}. 
The situation is often improved in practice by obtaining more images (e.g., SfM~\cite{schonberger2016structure} and SLAM~\cite{mur2015orb}),
or sensors like IMUs~\cite{guan2018minimal,ham2014hand,kneip2011robust} and RGBD~\cite{el2021unsupervisedr,yang2020extreme}. 
Nonetheless, people routinely infer relative pose from two ordinary images with a wide baseline, and whole industries like real estate 
depend on this ability. Rather then use extra sensors or images,
humans integrate cues like correspondence, 
familiar object size, and priors on scenes. This paper investigates such an ability, to
estimate relative pose, including rotation and translation with scale, from two ordinary images.

Based on these observations, there has been much work applying learning to the problem.
One line of
attack~\cite{detone2018superpoint,Dusmanu2019CVPR,sarlin2020superglue,sun2021loftr,tyszkiewicz2020disk}
has been to follow the classic pipeline and replace 
classic correspondence methods~\cite{bay2006surf,lowe2004distinctive,rublee2011orb} with learned ones. This
approach is appealing since the learning method finds correspondence, an especially
thorny challenge in the wide-baseline setting, and the conversion
of correspondences to pose is done by a provably correct method~\cite{longuet1981computer,nister2004efficient}. 
However, it comes at a cost of inheriting the Essential Matrix's intrinsic scale ambiguity,
leading to translation-up-to-scale.
Thus, another
line of work treats relative camera pose estimation as a
learning problem~\cite{cai2021extreme,en2018rpnet,jin2021planar,qian2020associative3d}. These approaches
have shown promise in the wide-baseline setting, but often involve multiple stages~\cite{chen2021wide,jin2021planar}, are not as performant as
correspondence-based techniques in the settings we try~\cite{en2018rpnet,qian2020associative3d}, or do not recover a translation
scale~\cite{cai2021extreme,chen2021wide}. Moreover, since these methods
learn an end-to-end mapping from images to camera pose with few
inductive biases, they are often data hungry.

\input{figures/teaser.tex}

We propose a Vision Transformer (ViT)~\cite{dosovitskiy2020image,vaswani2017attention} 
approach that  estimates rotation and {\it translation with scale}
in one forward pass by integrating the problem's structure implicitly as an inductive bias.
We reconcile the
Eight-Point Algorithm~\cite{hartley1997defense,longuet1981computer} with
ViTs by 
showing that 
a ViT forward pass can be made close to~\cite{hartley1997defense,longuet1981computer}
by
three minor modifications: (1) bilinear attention~\cite{kim2018bilinear} 
instead of attention~\cite{vaswani2017attention}; (2) 
quadratic position encodings; and (3) 
dual-softmax~\cite{Rocco18,sun2021loftr,tyszkiewicz2020disk} instead of 
softmax.
These modifications are put in one module, the {\it Essential Matrix Module (EMM)},
that we place atop an otherwise ordinary ViT, as shown in Fig.~\ref{fig:teaser}. The EMM gives an inductive
bias by providing positional features that 
approximate a key step of~\cite{longuet1981computer}, visual features, and features that mix the two.

We attach the Essential Matrix Module to the end of an ordinary
ViT~\cite{dosovitskiy2020image} described in \S\ref{sec:approach}. We train and evaluate this ViT on multiple
relative camera pose estimation tasks and datasets as described in \S\ref{sec:experiments} and compare with the state of the art
for each on challenging datasets like Matterport3D~\cite{chang2017matterport3d}, InteriorNet~\cite{li2018interiornet}, and StreetLearn~\cite{mirowski2019streetlearn}.
Our experiments on rotation+translation (\S\ref{sec:exp_rotationtranslation}) and rotation (\S\ref{sec:exp_rotation}) 
demonstrate: (1) that our simple approach outperforms (or occasionally matches) multiple alternate networks 
including concatenation~\cite{en2018rpnet,qian2020associative3d} and correlation volume
methods~\cite{cai2021extreme,jin2021planar}, techniques based on feature correspondence~\cite{detone2018superpoint,lowe2004distinctive},
and techniques trained to optimize a full 3D reconstruction~\cite{jin2021planar}; (2) that each component of the modification is important,
as shown by extensive ablations; and (3) that the EMM improves data efficiency by substantially boosting performance in moderate data
regimes (\S\ref{sec:exp_smalldata}), suggesting that epipolar geometry is a good inductive bias.

%% file: figures/teaser.tex
\begin{figure}[t]
	\centering
			{\includegraphics[width=\linewidth]{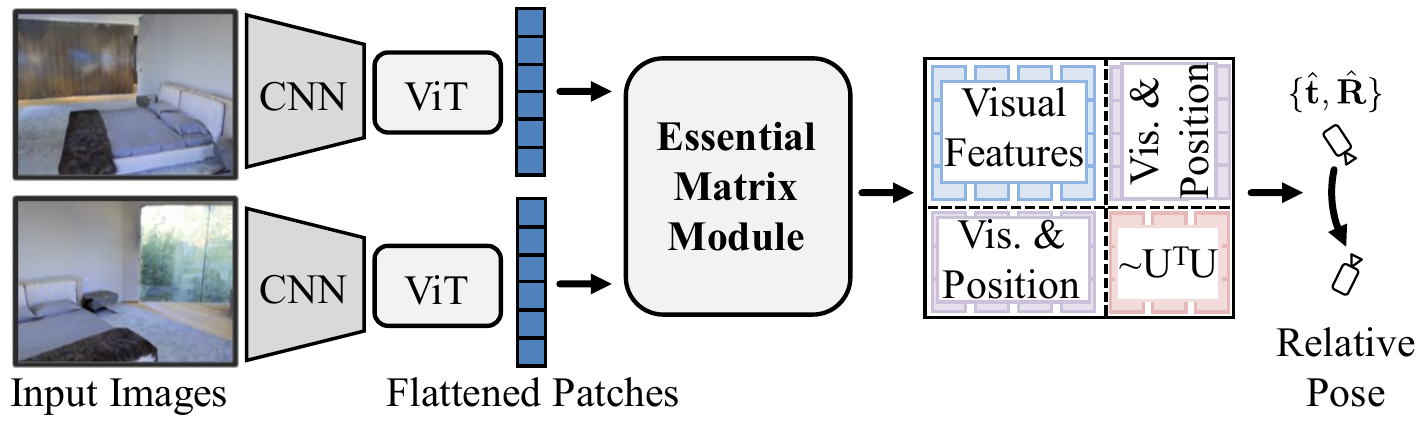}}
	\caption{We propose three small modifications to a ViT via the Essential Matrix Module, enabling computations similar to the Eight-Point algorithm. The resulting mix of visual and positional features is a good inductive bias for pose estimation.}
	\label{fig:teaser}
\end{figure}

%% file: related.tex
Our work introduces a learning-based approach to relative pose estimation by 
modifying vision transformers to perform computations similar to the Eight Point Algorithm.

\parnobf{Classic Work} Relative pose estimation from an
image pair is a sufficiently broad problem to preclude a full account.
We refer readers to~\cite{hartley2003multiple}, and focus on the closest works,
which all follow a strategy of solving for pose given correspondences
from local descriptors~\cite{bay2006surf,lowe2004distinctive,rublee2011orb}.
We revisit the 8-point algorithm~\cite{hartley1997defense,longuet1981computer} that maps correspondences to an Essential Matrix,
which was invented by Longuet-Higgins and 
extended to Fundamental Matrices by~\cite{faugeras1992can,hartley1992estimation}.
While it often replaced by other approaches that use fewer correspondences
(e.g.,~\cite{larsson2018beyond,nister2004efficient}), much of the 8-point algorithm's
structure are calculations that we show can be done by Transformers. 
In our wide baseline setting (i.e., large pose difference), historically
there are alternate descriptors and specialized techniques~\cite{matas2004robust,mishkin2015wxbs,Pritchett98a}.

\parnobf{Learned Pose Estimation} Given the difficulties associated
with optimization on correspondence, multiple lines of work aim to improve the pipeline with
learning. For instance, many methods improve
detectors and descriptors~\cite{detone2018superpoint,Dusmanu2019CVPR,tyszkiewicz2020disk} or correspondence estimation~\cite{brachmann2017dsac,probst2019unsupervised,ranftl2018deep,sarlin2020superglue,sun2021loftr,zhou2020learn}. 
These works typically 
turn correspondences to pose with the Essential Matrix~\cite{longuet1981computer,nister2004efficient},
which makes it impossible to 
recover translation scale without additional signals~\cite{hartley2003multiple}. In contrast, our proposed work learns a 
direct mapping without explicitly constructing an Essential Matrix, and therefore recovers scale using 
image-based cues. We note that our components are also often used in correspondence work~\cite{sun2021loftr,tyszkiewicz2020disk}; here,
we use them directly for pose and show a close relationship between ViTs and~\cite{longuet1981computer}.

Our method is closer to work that learns a mapping from images to pose. 
This area of research is relatively newer, and has become more complex over time (e.g., early networks concatenated data from two images~\cite{en2018rpnet,laskar2017camera,melekhov2017relative},
which has been supplanted by correlation volumes~\cite{cai2021extreme,jin2021planar}). 
These approaches are often data and compute hungry~\cite{chen2021wide}, use multiple stages (e.g., discrete/continuous optimization in~\cite{jin2021planar}, two-stage networks in~\cite{chen2021wide,wang2021tartanvo}), and use little of the structure of pose estimation. 
In contrast, our approach brings ViT computations close to this structure, which we hypothesize helps use the data more effectively.

\parnobf{SLAM, SfM, and RGBD} Given the difficulties of pose estimation from two images, a wealth of other approaches have been tried that modify the problem. 
The most common solution is to use more images with typically high overlap, e.g. via Structure-from-Motion~\cite{schonberger2016structure}, SLAM~\cite{mur2015orb,teed2021droid,wang2020tartanair} or localization~\cite{brahmbhatt2018geometry,radwan2018vlocnet,xue2019local}. 
In contrast, we aim to solve the two-view, wider-baseline case. 
Other solutions include adding sensors like an IMU~\cite{guan2018minimal,ham2014hand,kneip2011robust} or depth data~\cite{el2021unsupervisedr,yang2019extreme,yang2020extreme}; our approach relies only on RGB data.

\parnobf{Vision Transformers and Inductive Biases} Large parts of our proposed approach
follow a basic recipe for Vision Transformers~\cite{dosovitskiy2020image,vaswani2017attention}.
These have emerged as a competitor to convolutional neural networks in the past
few years, and we refer interested readers to~\cite{khan2021transformers,park2021vision} for a more thorough summary.
Our work shows that small modifications of the pipeline brings the computations
close those of~\cite{longuet1981computer}. This is part of a broader trend
of injecting geometric inductive biases to networks via layers~\cite{poursaeed2018deep}
or token engineering~\cite{yifan2021input}.

%% file: approach.tex
Our goal is to map two overlapping images to a relative camera pose {\it including translation scale}, or
a rotation $\RB \in \SOT$ and translation $\tB \in \mathbb{R}^3$.
This task requires both robustness to large view changes with limited correspondence, and handling scale ambiguity.
We propose a simple approach that fuses ideas from classical multi-view geometry with
large-scale learning.

At the heart of our approach is a transformer with small critical changes that 
mimic a computation used in the Eight Point Algorithm
~\cite{hartley1997defense,longuet1981computer}.
These changes include bilinear attention~\cite{kim2018bilinear},
dual-softmaxes~\cite{Rocco18,sun2021loftr,tyszkiewicz2020disk}, and an explicit positional encoding. 
We first analyze the relationship between the 
Eight Point Algorithm and
and an alternate setup that is more amenable to computation by a transformer (\S\ref{sec:approach_eightpt}).
We then describe how we operationalize this
by introducing our base transformer 
and our Essential Matrix Module (\S\ref{sec:approach_eightptinpractice}).
We conclude by analyzing the learnability of this function with synthetic experiments
(\S\ref{sec:approach_validation}).

\subsection{Transformers and the Eight Point Algorithm}
\label{sec:approach_eightpt}

The Fundamental and Essential matrices can be obtained from correspondences via the Eight-point algorithm~\cite{hartley1997defense,longuet1981computer,hartley1992estimation,faugeras1992can}.
As input, one assumes $N$ correspondences
$[u_i, v_i] \leftrightarrow [u'_i, v'_i]$. With known intrinsics $\KB$,
one represents the locations of the correspondences with normalized points 
$\xB_i \equiv \KB^{-1} [u_i, v_i, 1]^\top$ and $\xB'_i \equiv \KB^{-1} [u'_i, v'_i, 1]^\top$ and
recovers an Essential matrix ($\EB$); if $\KB$ is unknown, one uses standard homogeneous
coordinates (i.e., $\xB_i = [u_i, v_i, 1]^\top$)
and recovers a Fundamental matrix ($\FB$). Since we have the intrinsics, we
will refer to the Essential matrix.

The Eight-Point Algorithm constructs a matrix $\UB \in \mathbb{R}^{N \times 9}$ whose ith row $\UB_{i,:}$ is 
the Kronecker product of the correspondences, or $\xB_i \otimes \xB'_i$. The matrix
$\UB^\top \UB \in \mathbb{R}^{9\times 9}$ captures the information needed to estimate the Essential matrix:
one computes the eigenvector corresponding to the smallest
eigenvalue of $\UB^\top \UB$, reshapes the vector, and 
makes the reshaped matrix rank deficient. The resulting matrix $\EB$ does not uniquely define
the relative pose, but rather a family of solutions comprising two rotations $\RB$ and $\RB'$ and a translation direction
(that can be scaled by any $\lambda \not = 0$).

Careful minor modifications of Transformer can
enable the computation of the entries of $\UB^\top \UB$.
We assume the transformer is given a set of
$P$ patches at locations $\{\pB_j\}_{j=1}^{P}$ where every correspondence
is at one of the patches. In addition to using these locations directly, we
further define a 6D basis expansion $\phi([u,v,1]) = [1, u, v, uv, u^2, v^2]$ that we apply to each patch to yield a matrix $\PhiB \in \mathbb{R}^{P \times 6}$ such that $\PhiB_{j,:} = \phi(\pB_j)$.
Finally, to represent correspondences implicitly, we define an indicator matrix $\AB \in \{0,1\}^{P \times P}$
such that $\AB_{j,k} = 1$ if and only if points $\pB_k$ and $\pB_j$ are in
correspondence and 0 otherwise. 

Our key observation is that each unique entry of $\UB^\top \UB \in \mathbb{R}^{9 \times 9}$ is in 
the matrix $\PhiB^\top \AB \PhiB \in \mathbb{R}^{6 \times 6}$. While this more compact
form is not amenable to eigenvector analysis, it is all the information needed for a 
learned estimator.
A derivation appears in the supplement, but the two
critical steps are: first, to decompose the matrix as
an explicit sum over correspondences $\UB^\top \UB = \sum_{i=1}^N \UB_{i,:}^\top \UB_{i,:}$
and rewrite it implicitly with $\AB$; 
and second, that the 36 {\it unique} entries in 
$\UB_{i,:}^\top \UB_{i,:}$ can be generated from $\phi(\xB_i) \phi(\xB_i')^\top$.

The remaining step is estimation of $\RB$ and $\tB$ from $\UB^\top \UB$. 
MLPs are universal approximators~\cite{hornik1991approximation}, but a number of
things make this easier in practice. First, often one aims to solve
a subset of problems from a distribution, rather than 
{\it all} instances. Additionally, one is 
also using a wealth of alternate image-based cues. In addition
to facilitating learning, the network can use these cues to resolve the
ambiguities intrinsic to $\EB$: for instance, the scale
ambiguity can be resolved implicitly via recognizing familiar objects. We explore the learnability
of this function in \S\ref{sec:approach_validation}.

Together, this suggests that transformers estimating $\RB, \tB$ may benefit from a few small modifications.
The crux is that the computation of $\PhiB^\top \AB \PhiB$, using quadratic position encodings per patch
in $\PhiB$ and a correspondence indicator in $\AB$ mimics the computation of the entries of $\UB^\top \UB$. Thus,
a network may benefit from having $\PhiB^\top \AB \PhiB$ during prediction.  Moreover, $\AB$
also should be able to represent unmatched correspondences (i.e., $\sum_{k=1}^P \AB_{j,k}{\approx}0$).
Finally, we stress that the model should also contain features beyond $\PhiB$ 
to help learning and resolve ambiguities such as scale.

\input{figures/approach.tex}

\subsection{Putting things In Practice}
\label{sec:approach_eightptinpractice}

Our approach consists of two components. The main component is an 
{\it Essential Matrix Module}, which maps from $P$, $D$-dimensional transformer
tokens, one for each of the $P$ patches in the image, to a feature that is used to predict 
$\RB$ and $\tB$. This module is added to a standard
ViT~\cite{dosovitskiy2020image} backbone that maps images to a set of
tokens.  Our backbone deliberately follows a
standard vision transformer
recipe~\cite{dosovitskiy2020image,vaswani2017attention}: we see
backbone innovations as orthogonal to innovations in the mapping
from tokens to outputs. On the other hand, our Essential Matrix Module must
contain critical modifications.

\parnobf{Backbone and Setup} Our backbone consists of two main components 
that function as a learned mapping from an image to a $\mathbb{R}^{P \times D}$
matrix of features, one per patch. 
The first component is an encoder that
uses the first blocks from a standard ResNet-18~\cite{he2016deep}, which helps the network extract
good features per-patch. On top of this, we use blocks from a standard ViT~\cite{dosovitskiy2020image} (ViT-Tiny) to map the the
patch features to our final set of $P$ $D$-dimensional tokens. Since the architectures have different
feature sizes, we bridge them with a ResNet block that maps the feature dimensions. A full network
description appears in the supplemental.

\parnobf{Standard Transformer Model} The canonical ViT maps a set of patches
from one image to an output embedding used for classification. Given a patch
embedding, this entails computing query, key and value matrices $\QB, \KB, \VB
\in \mathbb{R}^{P \times D}$ followed by $\textrm{softmax}(\QB \KB^\top) \VB$. To avoid notational clutter, we drop the
usual~\cite{vaswani2017attention} scaling factor of $1/\sqrt{D}$ inside the
softmax here, and in all other softmax references.

For our case of two images, there are two sets of matrices, namely 
$\QB_1, \KB_1, \VB_1 \in \mathbb{R}^{P \times D}$ for image 1 and  $\QB_2, \KB_2, \VB_2 \in \mathbb{R}^{P \times D}$ for image 2. 
The simplest cross attention is to concatenate cross-attention per-image, or
\begin{equation}
\label{eqn:standardattention}
[\textrm{softmax}(\QB_1 \KB_2^\top)\VB_2, \textrm{softmax}(\QB_2 \KB_1^\top)\VB_1].
\end{equation}
This approach produces good results, but a few minor modifications can substantially improve its performance. 

\parnobf{Essential Matrix Module} We propose three changes to 
Eqn.~\ref{eqn:standardattention} that help approximate 
the entries of $\UB^\top \UB$.  These are shown in Fig.~\ref{fig:approach}.

\parnoit{Bilinear Attention and Quadratic Position Encodings} We apply bilinear attention~\cite{kim2018bilinear} 
to the values and quadratic positional encodings, or
\begin{equation}
\label{eqn:bilinearpool}
[\VB_2, \PhiB]^\top \textrm{norm}(\QB_1 \KB_2^\top) [\VB_2, \PhiB] \in \mathbb{R}^{(D+6) \times (D+6)}
\end{equation}
where $\PhiB \in \mathbb{R}^{P \times 6}$ contain the positional encodings $[1, u, v, uv, u^2, v^2]$ from
\S\ref{sec:approach_eightpt} and $\textrm{norm}$ is a normalization for the raw attention scores. Thus $\AB = \textrm{norm}(\QB_1 \KB_2^\top)$.
To use both images, we also compute Eqn.~\ref{eqn:bilinearpool} substituting in
$\QB_2$, $\KB_1$, and $\VB_1$ and concatenate the results, leading to a $(2D^2 + 24D + 72)$-dimensional feature per attention head.

\parnoit{Rationale} 
If $\AB = \textrm{norm}(\QB_1 \KB_2^\top)$ correctly indicates correspondence, 
then this computation makes the bottom-right $6 \times 6$ submatrix of Eqn.~\ref{eqn:bilinearpool} contain the entries of $\UB^\top \UB$. 
The top-left $D \times D$ submatrix
are visual features; the rest mix position and visual features. These image features are important for scale estimation, since
$\UB^\top \UB$ does not provide information about scale. They may also {\it
implicitly} contain position encodings (e.g., due to 
convolutions using zero-padding as a proxy for image location).
In practice, $\UB^\top \UB$ is followed by neural layers and thus does not need to match true $\UB^\top \UB$; though in the supplement we find non-zero rank correlation with ground truth.

\parnoit{Dual-softmax} The above is exact when $\AB$ is a correspondence indicator matrix. While 
attention makes this impossible to ensure exactly,  we help more closely
approximate it with a
dual-softmax~\cite{Rocco18,sun2021loftr,tyszkiewicz2020disk}, or
set $\textrm{norm}(\QB_1 \KB_2^\top)$ to
\begin{equation}
\textrm{softmax}(\QB_1 \KB_2^\top,1) \odot \textrm{softmax}(\QB_1 \KB_2^\top,2),
\end{equation}
where $\textrm{softmax}(\cdot,k)$ applies softmax across the k-th axis.

\parnoit{Rationale} Traditional attention normalizes the matrix product $\QB_1 \KB_2^\top \in \mathbb{R}^{P \times P}$ 
by a single softmax, $\textrm{softmax}(\QB_1 \KB_2^\top,1)$, 
forcing $\sum_{k=1}^P \AB_{j,k} = 1$. 
This constraint means that $\AB$ cannot indicate correspondence for patches without matches 
where $\sum_{k=1}^P \AB_{j,k} = 0$. At best, attention can be a uniform distribution; at worst, attention can latch onto a random
correspondence. In all cases, all patches contribute equally to the final product in Eqn.~\ref{eqn:bilinearpool}.
The network can mitigate this by making non-matching attention uniformly distributed and $\frac{1}{P} \sum_{k=1}^P \VB_{k,:} = 0$,
but this strategy does not work for parts of $\PhiB$, e.g., the $u^2$ term is non-negative and usually positive.

A dual-softmax suppresses non-matching patches while
not altering bidirectional matches. 
If the attention to-and-from patch $j$ is uniformly distributed, then the total attention $\sum_{k=1}^P \AB_{j,k}$ is $\frac{1}{P}$
instead of $1$ in the normal softmax case. On the other hand, if patch $j$ and patch $k$ both match well,
then the attention approaches $1$. Then, even if all but one patches have no match, their
total contribution is smaller ($\frac{P-1}{P}$) than even a single bidirectionally matching patch ($1$). 
Thus, the varying weighting helps suppress the contributions of patches without matches. 
While the form of $\AB$ intrinsically makes the computation an approximation, we stress that 
the consumer of $\PhiB \AB \PhiB^\top$ is a learned module and may be able to learn around
approximation errors, especially with vision features.

\parnobf{Pose Regressor} Given essential matrix encodings, we regress pose using a 2 hidden layer MLP.
We predict translation in real units, and predict rotation in quaternions, normalizing so scale is one.
We train only using a l1 geodesic loss on pose where the
geodesic loss is the magnitude of the vector between predicted and ground truth pose.

\subsection{Synthetic Validation}
\label{sec:approach_validation}

While our approximation of $\UB^\top \UB$ can be understood analytically, one critical
component is the learned mapping from $\UB^\top \UB$ to $\RB$, $\tB$ that is
done by the {\it Pose Regressor}.
To better understand the learnability of the function, we show the method on
synthetic examples with the entries of $\UB^\top \UB$ but no visual features.
Our scenes consist of points uniformly sampled inside a sphere with center
${\sim}\textrm{Unif}(-\frac{1}{2},\frac{1}{2})$ and radius ${\sim}\textrm{Unif}(\frac{1}{2},\frac{3}{2})$.
We sample camera rotations and translations from distributions that we vary 
to analyze the learnability of the problem.  For each pair of views with sufficient
overlap (100 of 10K sampled 3D points projecting to the images), we compute 
$\UB^\top \UB$, which is used as a feature for pose estimation by a MLP (details in supplement).

We analyze two tasks. The first is {\it Translation}, or estimating the generating $\tB$;
due to scale-ambiguity, we assume $||\tB||_2 = 1$ and $\tB_z > 0$.  We quantify errors by the angle between the estimated and true translation.
The second is {\it Rotation}, or estimating the rotation that generated the data, which 
forces the network to resolve the usual rotation ambiguity of $\EB$.
We quantify errors by the rotation geodesic. 

We try four distributions. In {\it 3D} $\RB$ is sampled via uniformly distributed Euler angles, and $\tB  \sim \textrm{Unif}(-1,1)$.
The next three are {\it 2D Small/Medium/Large}, consisting of 2D motion primarily in the xz plane with varying amounts of rotation variance: $\RB$ is sampled from Normally distributed Euler angles with 
rotation mainly in y ($y \sim N(0,r)$) and $x,z \sim N(0,\frac{r}{20})$) where $r = 1, 5, 25^\circ$ for small, medium, and large. Translation 
is mainly in the xz plane $\tB \sim N(0,[\frac{1}{3},\frac{1}{60},\frac{1}{3}])$.
To avoid epipolar degeneracies with no translation, we require $||\tB|| \ge \frac{1}{2}$.

We report results in Table~\ref{tab:utudemo} for models trained on 100K samples
using only $\UB^\top \UB$ as features, comparing
to chance for context. We compare models trained on 100K samples.
Even when trained on 100K samples and estimating
a general problem case, the networks learn the function. Once the data
is even moderately constrained (2D Large), relative errors drop considerably. This
suggests that the function is especially learnable under more constrained
rotations.

\begin{table}[t]
\caption{\textbf{Synthetic Validation.} We report the median angular error across two tasks (translation \& rotation)
and four datasets (in decreasing difficulty: 3D, 2D Large/Medium/Small (L/M/S).}
\label{tab:utudemo}
\centering
\begin{tabular}{@{}c@{~~}c@{~~}c@{~~}c@{~~}c@{~~}c@{~~}c@{~~}c@{~~}c@{}}
\toprule
 & \multicolumn{4}{c}{Translation (${}^\circ$)} & \multicolumn{4}{c}{Rotation (${}^\circ$)} \\
 & 3D & 2DL & 2DM & 2DS & 3D & 2DL & 2DM & 2DS \ \\
 \midrule
MLP & 18.4  & 5.6  & 3.0  & 1.8  & 33.5  & 3.6  & 1.8  & 0.7 \\
Chance & 64.0  & 49.1  & 47.9  & 47.9   & 125.3  & 22.2  & 4.8  & 1.0 \\
\bottomrule
\end{tabular}
\end{table}

\subsection{Implementation Details}

Full implementation details appear in the supplemental, and we will release code for reproducibility.
Our encoder is a pretrained ResNet-18, which we truncate to only use the first two of four modules, producing a $24 \times 24 \times 128$ feature map; we use an additional Residual block to map to feature size of 192 for the ViT.
We use the Timm~\cite{rw2019timm} ViT implementation, and use ViT-Tiny with a truncated depth of 5 plus our Essential Matrix Module. 
Outside of the proposed changes, our Essential Matrix Module follows a standard Cross-Attention Transformer Block architecture and normalization~\cite{lu2019vilbert}.
Positional encoding locations utilize known intrinsics, $\xB'_i \equiv \KB^{-1} [u'_i, v'_i, 1]^\top$, a manner similar to ~\cite{wang2021tartanvo}.
Each head of the Essential Matrix Module produces a $64$-D feature, which is $70$-D after concatenating the position encodings. With 3 heads, and the bilinear attention done once per image, 
this results in $3 \times 70^2 \times 2 = 29$K features.
We map this large feature to a hidden size of 512 for two hidden layers in our MLP before regressing 7D pose.
We implement using PyTorch~\cite{paszke2019pytorch} and use the LieTorch~\cite{teed2021tangent} extension for backpropogation of geodesic losses on quaternions.
We use learning rate of 5e-4 and train using Adam~\cite{kingma2014adam} optimizer and 1cycle scheduler~\cite{smith2019super} for 120k iterations with a batch size of 60 split over 10 GTX 1080 Tis, which takes about 1 day.

%% file: figures/approach.tex
\begin{figure*}[t]
	\centering
			{\includegraphics[width=\linewidth]{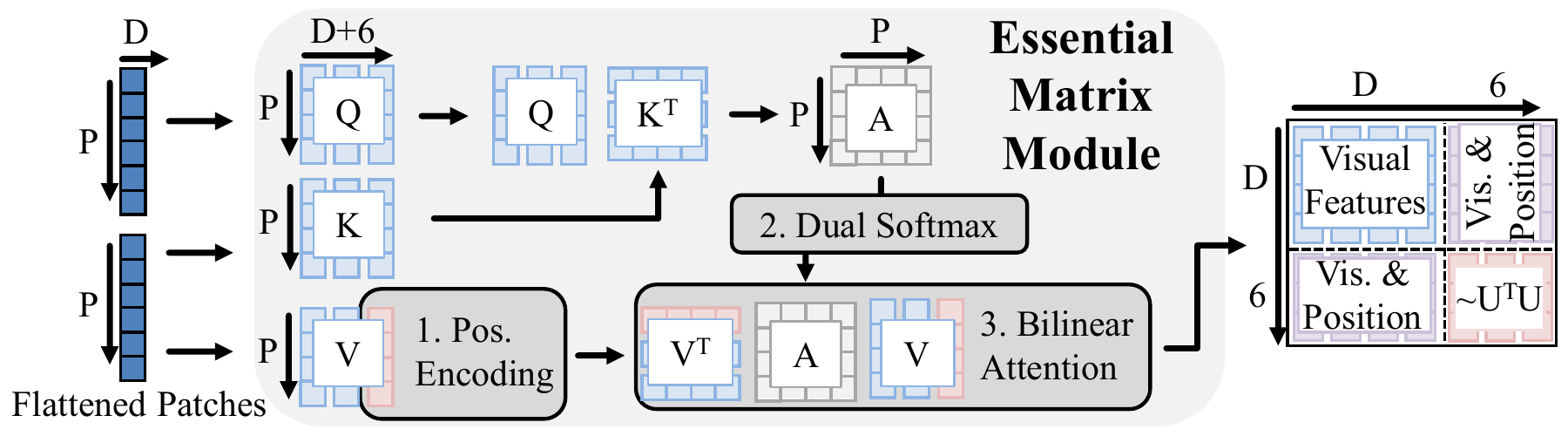}}
	\caption{\textbf{Essential Matrix Module.} We make three small changes to standard ViT Cross-Attention: (1) appending positional encodings to Values, (2) applying a dual softmax on Affinities, and (3) applying bilinear attention.}
	\label{fig:approach}
\end{figure*}

%% file: experiments.tex
We now evaluate the proposed method's ability to estimate relative pose in
comparison to the state of the art in two settings that share common metrics and evaluation 
settings~(\S\ref{sec:exp_metrics}). Our first task (\S\ref{sec:exp_rotationtranslation})
is wide baseline rotation {\it and translation} estimation, or estimating
a rotation in $\SOT$ and translation in $\mathbb{R}^3$ (i.e., including a scale).
The second task  (\S\ref{sec:exp_rotation}) is wide baseline rotation, or
estimating a rotation in $\SOT$ but
no translation. Finally, a crucial argument for our approach is that the modifications
of the transformer architecture serve as an inductive bias for the network. We
examine this empirically with experiments on substantially reduced data that test data
efficiency~(\S\ref{sec:exp_smalldata}).

\subsection{Metrics and Evaluation}
\label{sec:exp_metrics}

For each method, we compute the rotation error (defined as the rotation geodesic to the ground-truth)
and translation error (defined as the usual Euclidean distance to the ground truth), and
aggregate three summary statistics: the {\it mean}, the {\it median}, and the {\it percent of errors within a
threshold} that is task-specific (e.g, $30^\circ$) and will be described with each dataset. These capture different aspects
of the problem. Specifically, due to symmetries in the data, pose estimation errors are
often not unimodally distributed. Instead, often many results are highly accurate and 
a few are wrong by $90^\circ$ or $180^\circ$. The median error captures what a
typical prediction error is like and is outlier robust; the mean is the straight
average and is therefore sensitive to outliers; the percent within a threshold captures
a sense of how many predictions are ``reasonable'' for some threshold.

\subsection{Wide Baseline Rotation and Translation}
\label{sec:exp_rotationtranslation}

We begin by evaluating on our full problem, namely estimating a rotation in $\SOT$ and translation, {\it including scale}, in $\mathbb{R}^3$.
We follow the setup of~\cite{jin2021planar} to enable comparison with a variety of existing work and published baselines.

\parnobf{Dataset} We use data from Matterport3D~\cite{chang2017matterport3d} consisting of pairs of images with limited overlap (mean 2.3m translation, 53$^{\circ}$ rotation).
This dataset is a re-rendering of a real capture, using the Habitat~\cite{savva2019habitat} system. The train/val/test set of the dataset consist of 32K/5K/8K image pairs, respectively.
Following~\cite{jin2021planar}, we set the threshold for percent within a threshold to $30^\circ$ for rotation and $1$m for translation.

\parnobf{Baselines and Ablations}
Our primary comparison is the Sparse Planes method of~\cite{jin2021planar}, a strong baseline estimating both rotation and translation (including scale). Sparse Planes does joint reconstruction and pose estimation and consists of: initial reconstruction and camera estimation, discrete optimization, and a bundle-adjustment on SIFT features~\cite{lowe2004distinctive} extracted from texture that has been made fronto-parallel. 
The final step adds substantial complexity, so we compare to ({\it SparsePlanes~\cite{jin2021planar} No Bundle}) as well, which omits the final bundle adjustment, but still requires optimization. 
We also compare to the standalone pose estimation branch as {\it (Sparse Planes~\cite{jin2021planar} Camera Branch)}.
In addition, we compare to concurrent work ~\cite{agarwala2022planes} which closely builds off of Sparse Planes.

\input{tables/matterport.tex}

\input{tables/ess_ablations.tex}

We next report three baselines used by~\cite{jin2021planar}. 
The first is ({\it Associative 3D \cite{qian2020associative3d} camera branch}), which is an improved version of RPNet~\cite{en2018rpnet}. 
The second is the reconstruction-based RGBD odometry method of Raposo et al.~\cite{raposo2013plane} applied to~\cite{ranftl2020towards}.
Third, we compare with ({\it SuperGlue~\cite{sarlin2020superglue}}), using the settings from~\cite{jin2021planar}. 
In addition, we compare to LoFTR~\cite{sun2021loftr}. Like SuperGlue, LoFTR supervises correspondences, and therefore requires depth supervision in addition to pose, and cannot recover translation scale.

Finally, we compare with four ablations that 
test the contributions of our method. All methods use the same MLP Regressor, and
full descriptions of these appear in the supplement. The first is ({\it CNN Pose Regressor}),
which predicts pose from concatenated base CNN extracted features. This
gives a sense of how a simple method does.
The second is ({\it +ViT}), which adds a ViT that is capped with standard attention (Eqn.~\ref{eqn:standardattention})
on top of the backbone. This tests the contribution of a ViT {\it without} the Essential Matrix Module.
The third is ({\it +Bilinear}) which replaces standard attention (Eqn.~\ref{eqn:standardattention})
with bilinear attention, but without dual-softmax and quadratic positional encodings. Finally, we report
({\it +Dual Softmax}), which adds dual softmax. 

\input{tables/interior_streetlearn.tex}

\input{figures/qualitative.tex}

\begin{figure}
\centering
\begin{tabular}{@{}c@{}c@{}}
\includegraphics[width=0.475\linewidth]{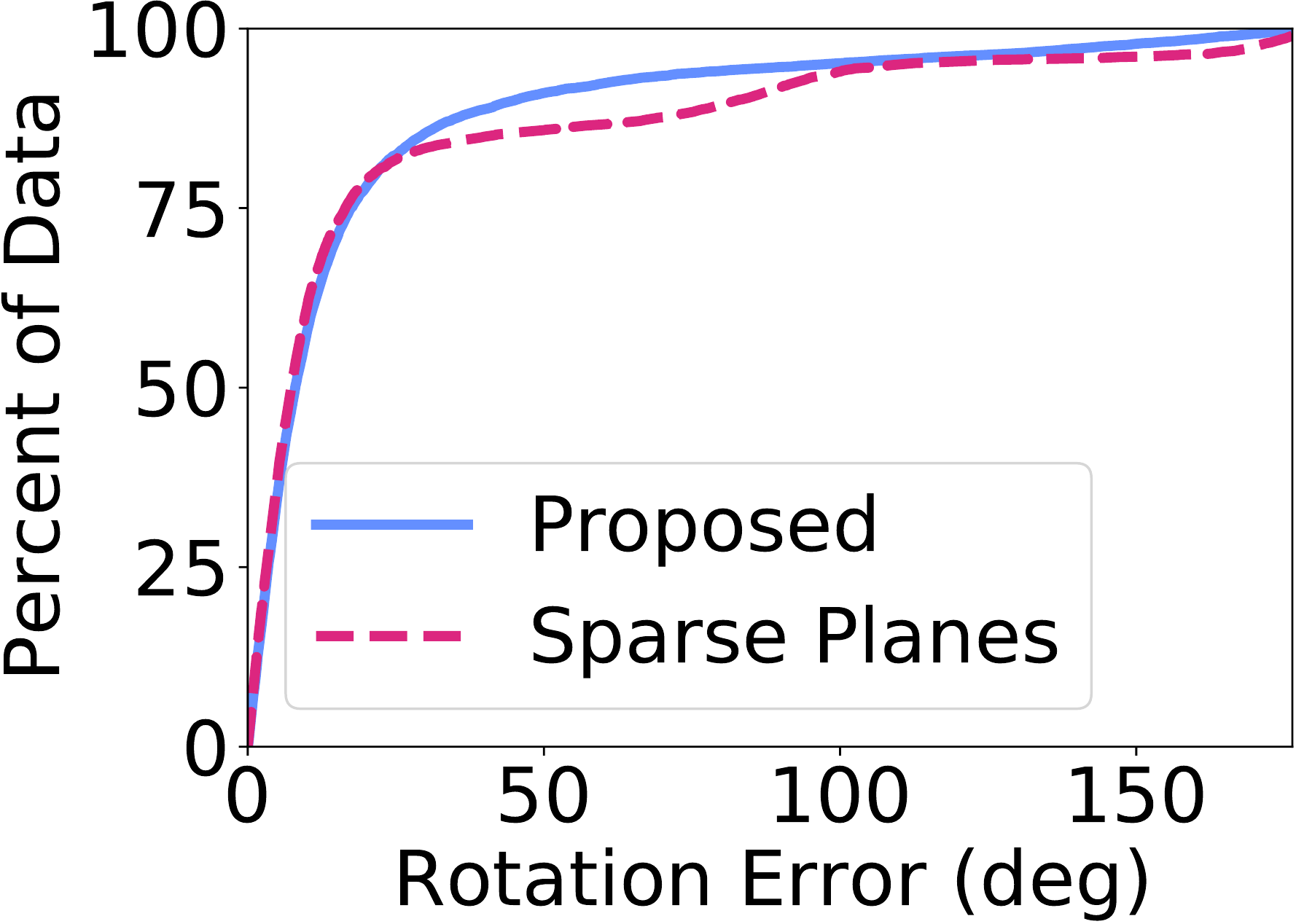} &
\includegraphics[width=0.475\linewidth]{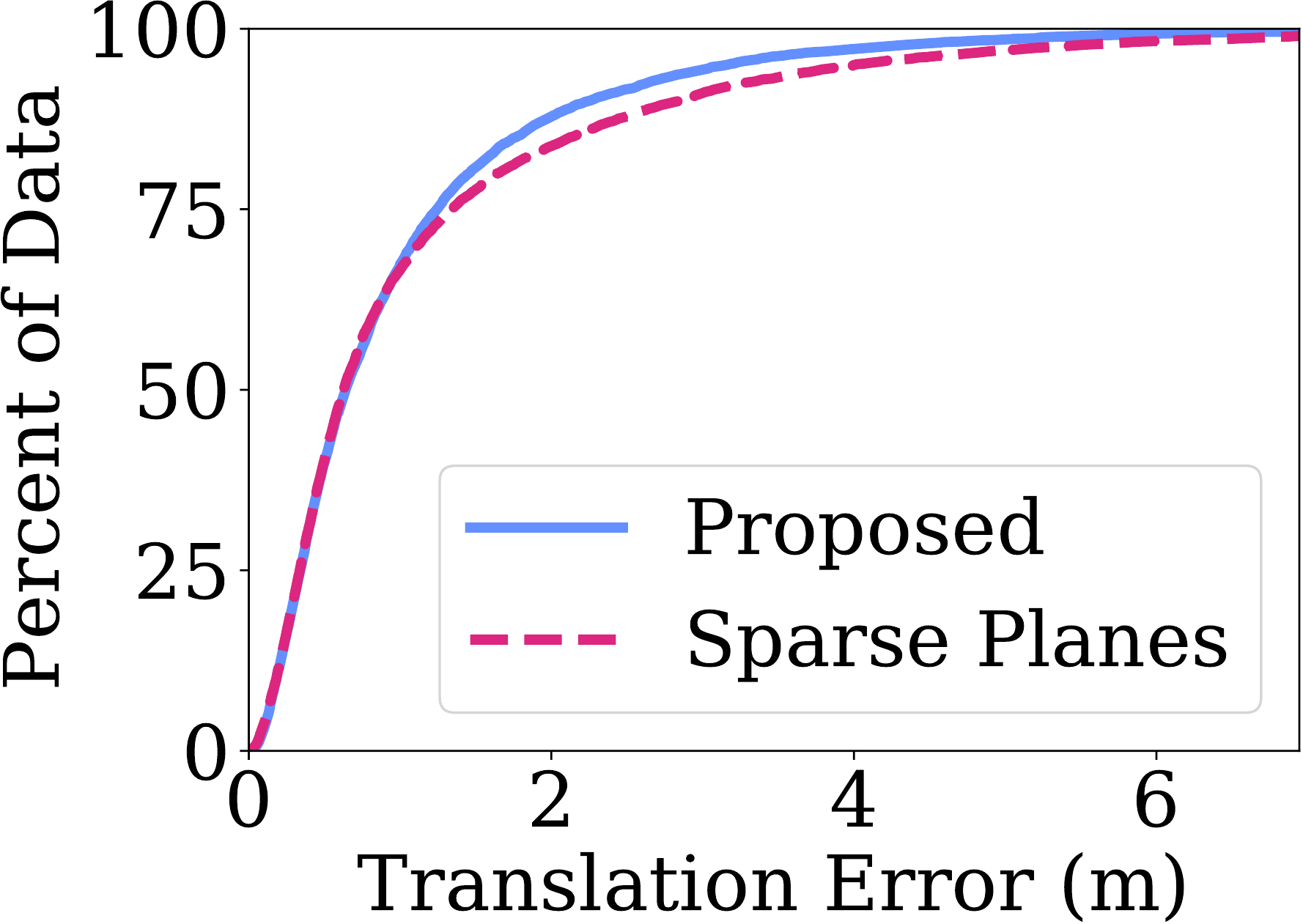} \\
\end{tabular}
\caption{\textbf{Error CDFs on Matterport}. The proposed approach
shows increased robustness to large view changes.}
\label{fig:cdf_matterport}
\end{figure}

\parnobf{Quantitative Results}
We report results in Table ~\ref{tab:matterport}. Joint prediction of rotation and translation (including scale) on wide-baseline pairs is a challenging problem. Non-trivial methods \cite{qian2020associative3d,ranftl2020towards,raposo2013plane} have less than 40\% of predictions within 30$^\circ$ of true rotation, and less than 20\% of errors within 1m of true translation. 
Methods that are most competitive (LoFTR~\cite{sun2021loftr}, SuperGlue \cite{sarlin2020superglue}, \cite{jin2021planar}) require depth supervision in addition to pose, while the best rotation results (\cite{sun2021loftr}, \cite{sarlin2020superglue}) are produced by correspondence-based methods not predicting translation scale. 
Of methods producing translation scale, ours typically performs best, and it outperforms SuperGlue in both average rotation and percentage within 30$^\circ$.

Ablations, shown in Table~\ref{tab:ess_ablations}, show the reasons for success. As with \cite{qian2020associative3d,ranftl2020towards,raposo2013plane}, CNN and ViT models struggle at the task. 
Adding bilinear attention reduces errors tremendously -- reducing median rotation error by two thirds -- while the dual softmax reduces errors further significantly.
Adding positional encodings further improves performance across all measurements.

\parnobf{Analysis} Qualitative results, in Figure \ref{fig:qualitative}, are
consistent with quantitative findings. Namely, predicted pose more closely
matches ground truth on difficult examples, resulting in much better mean
performance than baselines. Error vs. view change is analyzed further in the
Supplemental. Figure~\ref{fig:cdf_matterport} displays error CDFs on Matterpor
Compared to the most competitive baseline
\cite{jin2021planar} (Full), the proposed method has fewer very
large errors.

\subsection{Wide Baseline Rotation}
\label{sec:exp_rotation}

\input{tables/interior_streetlearn_ablations.tex}

We next study wide baseline rotation, where we compare with~\cite{cai2021extreme} and their baselines.

\parnobf{Datasets} We use the two datasets from~\cite{cai2021extreme}, which were derived from panoramic photos 
and follow the setup of~\cite{cai2021extreme}. The first dataset is
InteriorNet~\cite{li2018interiornet}, which consists of 10,050 panoramic views across 112 synthetic houses. Of these, 82 houses
are allocated for training and the remaining 30 houses are used for testing. The dataset has 610k image pairs (350K overlapping), with a test set of 1K pairs.
StreetLearn~\cite{mirowski2019streetlearn}, consists of panoramic outdoor images in New York City
that have been scrubbed to ensure privacy (full details in Supplemental). This dataset has
1.1M train pairs (460K with overlap), and 1K test pairs from a set of 143K panoramas.  
We additionally evaluate on the ``InteriorNet-T'' and ``StreetLearn-T'' datasets, which select from {\it different} panoramas for each image in a pair, resulting in translation in addition to rotation.
This translation is not, however, estimated in this setting.
To facilitate comparisons we use $10^\circ$ as a threshold for rotation error following~\cite{cai2021extreme}.
We use the setup of Cai \textit{et al.}~\cite{cai2021extreme} using only overlapping images, and breaking down overlap into {\it large} overlap (less than 45$^\circ$ rotation) and {\it small} overlap (more than 45$^\circ$).
Cai \textit{et al.} also conduct experiments on non-overlapping images; we consider this beyond our scope, which is focused on the case where correspondences may exist.

\parnobf{Baselines and Ablations} 
We compare to the state of the art ({\it Extreme Rotation~\cite{cai2021extreme}}), which computes a cross-correlation volume on paired image features, and uses a CNN to classify pose. We also
report this method's baselines: Reg6D~\cite{zhou2019continuity}, which predicts a 6D representation from concatenated image features, similar to the {\it Associative3D Camera Branch} from \S\ref{sec:exp_rotationtranslation}
as well as correspondence baselines SIFT~\cite{lowe2004distinctive} and SuperPoint~\cite{detone2018superpoint}. These baselines occasionally fail. Following~\cite{cai2021extreme}, we indicate failure on 
more than 50\% of the test set by marking the number in gray. We report the same ablations as in \S\ref{sec:exp_rotationtranslation}.

\parnobf{Quantitative Results} The proposed method is typically better than all baselines across both InteriorNet and StreetLearn, for both versions and overlap settings of the dataset (Table ~\ref{tab:noah}).
Often, the proposed method reduces error compared to competing methods by more than half (e.g., InteriorNet Mean, Median with Large Overlap; StreetLearn-T Mean with Large Overlap).
Small overlap is an especially difficult setting. For instance, on InteriorNet-T, all baselines have mean error above 10$^\circ$. Yet, the proposed method is within 10$^\circ$ more than \textit{96\%} of the time.
Interestingly, median error on InteriorNet-T is worse than Cai \textit{et al.}~\cite{cai2021extreme}. 
We believe the large scale of InteriorNet is not the method's strongest setting, and the method provides strong inductive bias for small data settings (see \S\ref{sec:exp_smalldata}).
Nevertheless, we consider Cai \textit{et al.} to be a strong baseline as it is specialized to large angle changes.

Performance breakdown of the model is displayed in Table ~\ref{tab:noah2}. 
Adding a ViT is quite important, likely attributable to the large scale of data available.
Beyond the ViT, improvements by each step are more mixed compared to the clear improvement of each step on Matterport.
For instance, adding the dual softmax without coordinate embeddings is typically not helpful compared to using only Bilinear Attention.
Yet, the full model performs best (best 5 times, second best 3 times; Bilinear Attention is best 4 times, second best twice).
Moreover, the full model is rarely significantly worse than any intermediate ablation.
This suggests, as argued in \S\ref{sec:approach}, that all of the proposed components work together. 
We emphasize these settings have extraordinary numbers of views. 
\S\ref{sec:exp_smalldata} will show the substantially higher data efficiency of
the essential module.

\input{figures/qualitative2.tex}

\parnobf{Analysis} Qualitative results validate quantitative findings in Fig.~\ref{fig:qualitative2}. While the evaluation datasets have huge rotations across indoor and outdoor settings, the proposed model is typically accurate, often even within 1\% of true rotations.

\subsection{Effectiveness on Smaller Datasets}
\label{sec:exp_smalldata}

One of the primary arguments for the use of the proposed network structure is
that it provides a useful inductive bias by helping the network compute information that is known
to constrain the set of feasible rotations and translations. In principle, since feedforward networks
are universal approximators~\cite{hornik1991approximation}, networks ought to be able to learn to estimate
relative pose with enough data. However, the right inductive biases ought to let them learn {\it faster}.

We now examine performance as a function of number of images. First, this helps empirically assess
whether various networks structures provide useful inductive biases. Second, this is of practical concern
since it tests data efficiency.

\parnobf{Datasets} We use InteriorNet-T and StreetLearn-T from \S\ref{sec:exp_rotation}, with significantly reduced 32K train image pairs. 
Collecting large-scale datasets such as these is challenging without a simulator or specialized company resources, so this smaller scale may be more realistic for e.g. user-collected posed images.

\parnobf{Ablations and Results} Our primary comparison is with the ViT baseline. Because it is a near alternative to our proposed Essential Matrix Module, we can measure the impact of our main contributions. Results are presented in Table~\ref{tab:data}, which is a reduced version of Table~\ref{tab:noah2}, with results also on the 32K image train set. 
Across datasets, the proposed method scales significantly better to a small train set. Even in cases the ViT slightly outperformed our proposed full model with full set, the inductive biases of the proposed method give it substantial improvement in the small setting.

\begin{table}
\caption{\textbf{Performance with limited data}. The proposed method scales better to small data than a typical learned model (e.g. ViT), indicating better inductive biases.}
\label{tab:data}
\resizebox{\ifdim\width>\linewidth \linewidth \else \width \fi}{!}{
\begin{tabular}{lccccccc}
\toprule
 & & \multicolumn{6}{c}{InteriorNet-T}
\\
    & & \multicolumn{3}{c}{Full} & \multicolumn{3}{c}{32K} 
\\
    & & Avg & Med & $\%<10^\circ$ & Avg & Med & $\%<10^\circ$
\\ 
\midrule
\multirow{2}{*}{Large} & ViT & 0.61 & 0.49 & \textbf{100.00} & 5.78 & 3.23 & 92.84
\\
& Full & \textbf{0.48} & \textbf{0.40} & \textbf{100.00} & \textbf{4.44} & \textbf{2.58} & \textbf{95.82}
\\ 
\midrule
\multirow{2}{*}{Small} & ViT & \textbf{1.44} & 1.09 & \textbf{100.00} &  11.89 & 4.38 & 78.70
\\
& Full & 1.81 & \textbf{0.94} & \textbf{100.00} & \textbf{8.22} & \textbf{4.27} & \textbf{89.20}
\\ 

\bottomrule
~ &  ~
\\ \toprule
    & & \multicolumn{6}{c}{StreetLearn-T}
\\
   & & \multicolumn{3}{c}{Full} & \multicolumn{3}{c}{32K} 
\\
    & & Avg & Med & $\%<10^\circ$ & Avg & Med & $\%<10^\circ$
\\ \midrule
\multirow{2}{*}{Large} & ViT & \textbf{3.52} & 2.56 & \textbf{94.74} &  11.51 & 7.69 & 56.58
\\
& Full & 4.08 & \textbf{2.43} & 90.13 & \textbf{7.22} & \textbf{4.44} & \textbf{81.58}
\\ 
\midrule
\multirow{2}{*}{Small} & ViT & 12.93 & \textbf{3.16} & 84.46 & 29.28 & 14.94 & 36.59
\\
& Full & \textbf{9.19} & 3.25 & \textbf{87.70} & \textbf{13.29} & \textbf{5.55} & \textbf{71.72}
\\ 
\bottomrule
\end{tabular}
}
\end{table}

%% file: tables/matterport.tex
\begin{table}
    \caption{\textbf{Translation and Rotation Performance on Matterport.} Ours is best among methods producing translation scale. All baselines supervise depth except \cite{jin2021planar} (Camera Br) and Ours.}
\label{tab:matterport}

\resizebox{\ifdim\width>\columnwidth \columnwidth \else \width \fi}{!}{

\begin{tabular}{l c c c c c c} \toprule
  & \multicolumn{3}{c}{Translation (m)} & \multicolumn{3}{c}{Rotation (degrees)} \\
Method & Med.$\downarrow$ & Avg.$\downarrow$ & $\leq$1m$\uparrow$ & Med.$\downarrow$ & Avg.$\downarrow$ & $\leq$ 30$\uparrow$ \\
\midrule
\cite{raposo2013plane} + \cite{ranftl2020towards} & 3.34 & 4.00 & 8.3 & 50.98 & 57.92 & 29.9 \\
Assoc.3D~\cite{qian2020associative3d} & 2.17 & 2.50 & 14.8 & 42.09 & 52.97 & 38.1   \\
\cite{jin2021planar} (Camera Br) & 0.90 & 1.40 & 55.5 & 7.65 & 24.57 & 81.9 \\
\cite{jin2021planar} (No Bundle) & 0.88 & 1.36 & 56.5 & 7.58 & 22.84 & 83.7   \\
\cite{jin2021planar} (Full) & \textbf{0.63} & 1.25 & 66.6 & 7.33 & 22.78 & 83.4   \\
PlaneFormers \cite{agarwala2022planes} & 0.66 & 1.19 & 66.8 & 5.96 & 22.20 & 83.8   \\
Ours & 0.64 & \textbf{1.01} & \textbf{67.4} & 8.01 & 19.13 & 85.4 \\ 
SuperGlue~\cite{sarlin2020superglue} & - & - & - & 3.88 & 24.17 & 77.8   \\
LoFTR~\cite{sun2021loftr} & - & - & - & \textbf{0.71} & \textbf{11.11} & \textbf{90.5}   \\
\bottomrule
\end{tabular}
}
\end{table}

%% file: tables/ess_ablations.tex
\begin{table}
    \caption{\textbf{Essential Matrix Module Ablations on Matterport.} All three components of the Essential Matrix Module yield meaningful improvement across metrics.}
\label{tab:ess_ablations}

\resizebox{\ifdim\width>\columnwidth \columnwidth \else \width \fi}{!}{

\begin{tabular}{l c c c c c c} \toprule
  & \multicolumn{3}{c}{Translation (m)} & \multicolumn{3}{c}{Rotation (degrees)} \\
Method & Med.$\downarrow$ & Avg.$\downarrow$ & $\leq$1m$\uparrow$ & Med.$\downarrow$ & Avg.$\downarrow$ & $\leq$30$\uparrow$ \\
\midrule
CNN Pose Regressor & 1.53 & 1.83 & 28.6 & 31.31 & 45.05 & 48.8\\
+ViT & 1.47 & 1.79 & 30.1 & 29.9 & 43.33 & 50.1   \\
+Bilinear Attention & 1.13 & 1.49 & 44.5 & 9.76 & 28.36 & 73.1 \\
+Dual Softmax & 0.70 & 1.06 & 64.8 & 8.62 & 21.23 & 83.3   \\
Full & \textbf{0.64} & \textbf{1.01} & \textbf{67.4} & \textbf{8.01} & \textbf{19.13} & \textbf{85.4}   \\
\bottomrule
\end{tabular}
}
\end{table}

%% file: tables/interior_streetlearn.tex
\begin{table*}
    \caption{\textbf{Rotation Performance on InteriorNet and StreetLearn.} We train and evaluate on only overlapping images. ``*'' indicates the method sometimes failed to produce pose estimation; errors were calculated only on successful image pairs. Gray text indicates failure over 50\% of test pairs. The proposed method outperforms alternatives almost universally and often significantly.}
\label{tab:noah}
\resizebox{\ifdim\width>\linewidth \linewidth \else \width \fi}{!}{
\begin{tabular}{l l c c c c c c c c c c c c} \toprule
  & & \multicolumn{3}{c}{InteriorNet} & \multicolumn{3}{c}{InteriorNet-T} & \multicolumn{3}{c}{StreetLearn} & \multicolumn{3}{c}{StreetLearn-T} \\
Overlap & Method & Avg ($^{\circ}$ $\downarrow$) & Med. ($^{\circ}$ $\downarrow$) & 10 (\% $\uparrow$) & Avg ($^{\circ}$ $\downarrow$) & Med. ($^{\circ}$ $\downarrow$) & 10 (\% $\uparrow$) &  Avg ($^{\circ}$ $\downarrow$) & Med. ($^{\circ}$ $\downarrow$) & 10 (\% $\uparrow$) &  Avg ($^{\circ}$ $\downarrow$) & Med. ($^{\circ}$ $\downarrow$) & 10 (\% $\uparrow$) \\
\midrule
\multirow{5}{*}{Large} & SIFT*~\cite{lowe2004distinctive} & 6.09 & 4.00 & 84.86 & 7.78 & 2.95 & 55.52 & 5.84 & 3.16 & 91.18 & \textcolor{gray}{18.86} & \textcolor{gray}{3.13} & \textcolor{gray}{22.37} \\
& SuperPoint*~\cite{detone2018superpoint} & 5.40 & 3.53 & 87.10 & 5.46 & 2.79 & 65.97 & 6.23 & 3.61 & 91.18 & \textcolor{gray}{6.38} & \textcolor{gray}{1.79} & \textcolor{gray}{16.45} \\
& Reg6D~\cite{zhou2019continuity} & 5.43 & 3.87 & 87.10 & 10.45 & 6.91 & 67.76 & 3.36 & 2.71 & 97.65 & 12.31 & 6.02 & 69.08  \\
& Cai \emph{et al.}~\cite{cai2021extreme} & 1.53 & 1.10 & 99.26 & \textbf{2.89} & \textbf{1.10} & 97.61 & 1.19 & 1.02 & 99.41 & 9.12 & 2.91 & 87.50 \\
& Ours & \textbf{0.48} & \textbf{0.40} & \textbf{100.00} & 2.90 & 1.83 & \textbf{97.91} & \textbf{0.62} & \textbf{0.52} & \textbf{100.00} & \textbf{4.08} & \textbf{2.43} & \textbf{90.13} \\
\midrule
\multirow{5}{*}{Small} & SIFT*~\cite{lowe2004distinctive} & 24.18 & 8.57 & \textcolor{gray}{39.73} & \textcolor{gray}{18.16} & \textcolor{gray}{10.01} & 18.52 & 16.22 & 7.35 & 55.81 & \textcolor{gray}{38.78} & \textcolor{gray}{13.81} & \textcolor{gray}{5.68} \\
& SuperPoint*~\cite{detone2018superpoint} & \textcolor{gray}{16.72} & \textcolor{gray}{8.43} & \textcolor{gray}{21.58} & \textcolor{gray}{11.61} & \textcolor{gray}{5.82} & \textcolor{gray}{11.73} & \textcolor{gray}{19.29} & \textcolor{gray}{7.60} & \textcolor{gray}{24.58} & \textcolor{gray}{6.80} & \textcolor{gray}{6.85} & \textcolor{gray}{0.95} \\
& Reg6D~\cite{zhou2019continuity} &  17.83 & 9.61 & 51.37 & 21.87 & 11.43 & 44.14 & 7.95 & 4.34 & 87.71 & 15.07 & 7.59 & 63.41 \\
& Cai \emph{et al.}~\cite{cai2021extreme} & 6.45 & 1.61 & 95.89 & 10.24 & \textbf{1.38} & 89.81 & 2.32 & 1.41 & 98.67 & 13.04 & 3.49 & 84.23 \\
& Ours & \textbf{1.81} & \textbf{0.94} & \textbf{99.32} & \textbf{4.48} & 2.38 & \textbf{96.30} & \textbf{1.46} & \textbf{1.09} & \textbf{100.00} & \textbf{9.19} & \textbf{3.25} & \textbf{87.70} \\
\bottomrule
\end{tabular}
}
\end{table*}

%% file: figures/qualitative.tex
\begin{figure}[t]
	\centering
			{\includegraphics[width=\linewidth]{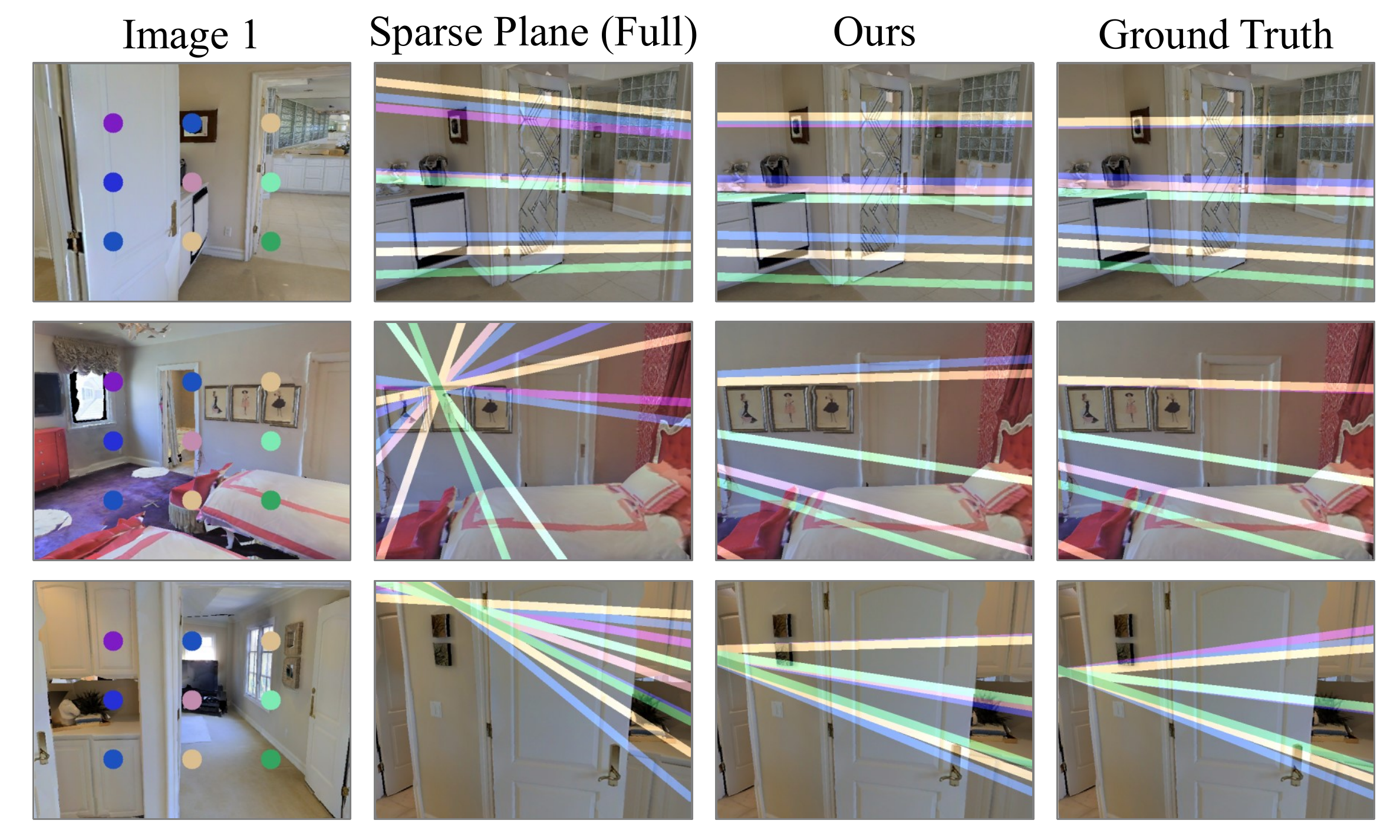}}
	\caption{\textbf{Epipolar Lines on Matterport.} Our predictions better match true pose, particularly on large view changes.}
	\label{fig:qualitative}
\end{figure}

%% file: tables/interior_streetlearn_ablations.tex
\begin{table}
    \caption{\textbf{Rotation Ablations InteriorNet and StreetLearn.} (Second best underlined). The ViT significantly improves over CNN only. Components of the proposed model perform in different settings, but the full model is often best and typically competitive with the best ablation, while ablations sometimes do poorly (Bilinear Att. on InteriorNet Small, ViT on StreetLearn Small).}
\label{tab:noah2}
\resizebox{\ifdim\width>\linewidth \linewidth \else \width \fi}{!}{
\begin{tabular}{l l c c c c c c} \toprule
  & & \multicolumn{3}{c}{InteriorNet-T} & \multicolumn{3}{c}{StreetLearn-T} \\
Overlap & Method & Avg ($^{\circ}$ $\downarrow$) & Med. ($^{\circ}$ $\downarrow$) & 10 (\% $\uparrow$) & Avg ($^{\circ}$ $\downarrow$) & Med. ($^{\circ}$ $\downarrow$) & 10 (\% $\uparrow$) \\
\midrule
\multirow{5}{*}{Large} & CNN Pose Regressor & 5.29 & 2.6 & 89.85 & 15.25 & 10.00 & 50.00 \\
& +ViT & \underline{2.99} & 1.64 & 96.72 & \textbf{3.52} & 2.56 & \textbf{94.74} \\
& +Bilinear Attention & 3.25 & \textbf{1.49} & \underline{97.31} & 4.73 & 2.68 & \underline{92.76} \\
& +Dual Softmax & 6.03 & \underline{1.63} & 93.43 & 4.39 & 2.64 & 91.45 \\
& Full & \textbf{2.90} & 1.83 & \textbf{97.91} & \underline{4.08} & \textbf{2.43} & 90.13 \\
\midrule
\multirow{5}{*}{Small} & CNN Pose Regressor & 19.79 & 4.05 & 69.44 & 29.95 & 15.22 & 34.07 \\
& +ViT & \underline{5.43} & 2.00 & \underline{94.75} & 12.93 & \textbf{3.16} & 84.86 \\
& +Bilinear Attention & 8.54 & \textbf{1.79} & 90.43 & \textbf{8.70} & 3.41 & \textbf{89.59} \\
& +Dual Softmax & 10.44 & \underline{1.96} &	89.51 & 10.74 & \underline{3.24} & 87.07 \\
& Full & \textbf{4.48} & 2.38 & \textbf{96.30} & \underline{9.19} & 3.25 & \underline{87.70} \\
\bottomrule
\end{tabular}
}
\end{table}

%% file: figures/qualitative2.tex
\begin{figure}[t]
	\centering
			{\includegraphics[width=\linewidth]{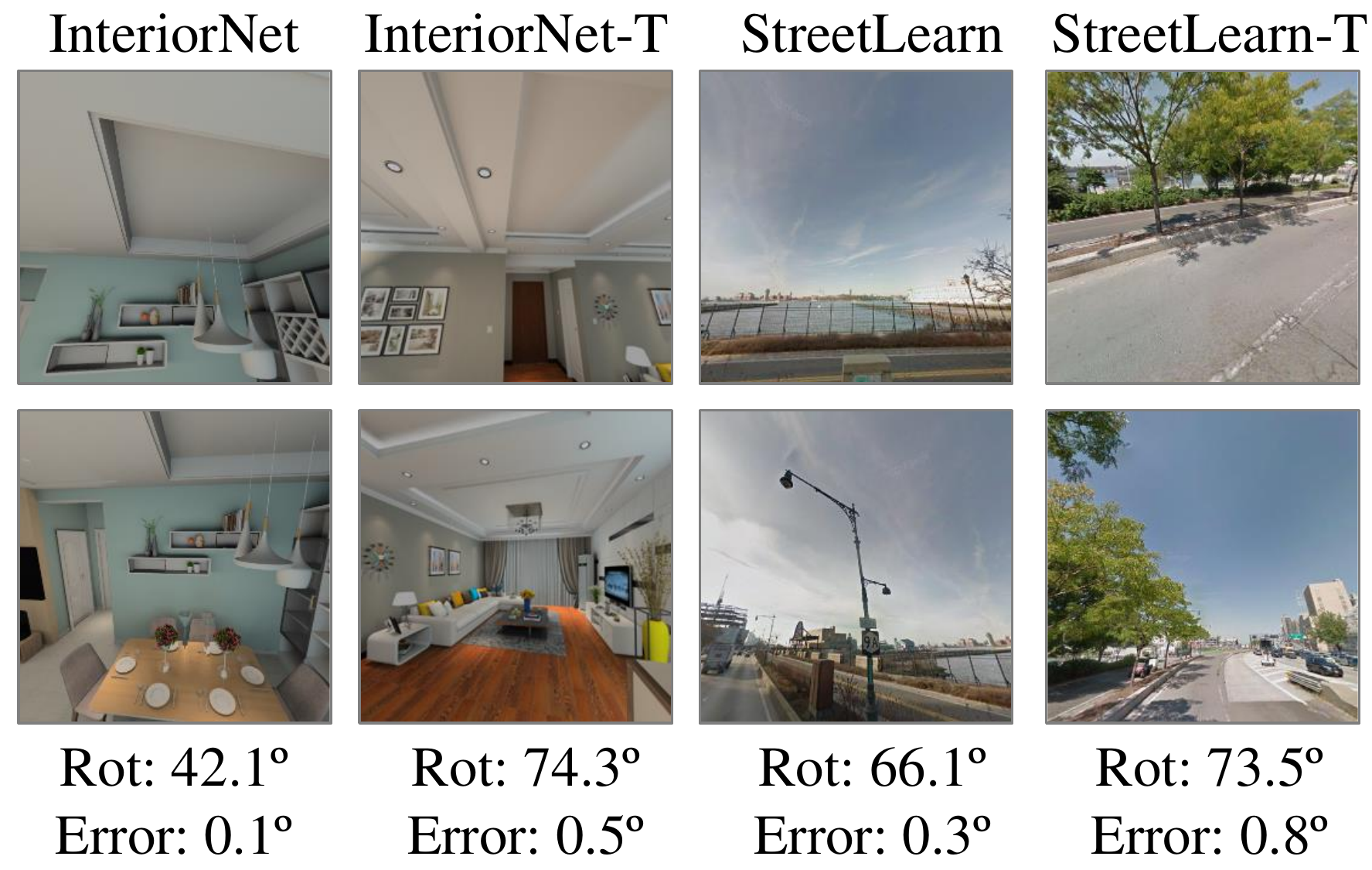}}
	\caption{\textbf{Error vs. Rotation.} The proposed method produces high precision when faced with large view change.}
	\label{fig:qualitative2}
\end{figure}

%% file: discussion.tex
In this paper we presented a simple and interpretable end-to-end approach for pose estimation.
Our key technical contribution is to implicitly represent correspondences from a ViT using an essential matrix module, from which an MLP can estimate pose.
Theoretical results show this formulation can approximate the matrix $\UB^\top \UB$ that is analyzed in the Eight Point algorithm; empirical results show given this, the MLP can suitably estimate pose.
While alternatives make additional assumptions about input or require optimization, this method requires only paired RGB images as input, and is competitive in a variety of settings and viewpoint changes while being computationally efficient.

\parnobf{Limitations and Social Impact} The model is generally robust across view change. 
However, other methods are better suited for the two extremes in view change. 
In the case of small view change, the transformer is limited in terms of precision by the number of patches. 
Alternative CNN-based methods such as~\cite{wang2021tartanvo} may more easily operate upon high resolution.
The model is also not prepared to predict pose on images with no overlap or correspondences; classification-based work e.g.~\cite{cai2021extreme} is better suited for this. 
Using datasets such as Matterport collected in nice homes leads to models which will likely perform better in these homes and possibly not as well in less expensive homes. 
Using synthetic data such as InteriorNet may help combat this bias. 
Training and evaluating on StreetView images should be handled with special care, as these images can contain personal information. 
The original authors blurred faces in the dataset, and a random manual search of 500 images also revealed no personal identifying information. 

\parnobf{Acknowledgments} Thanks to Linyi Jin, Ruojin Cai and Zach Teed for help replicating and building upon their works. 
Thanks to Mohamed El Banani, Karan Desai and Nilesh Kulkarni for their helpful suggestions. 
Thanks to Laura Fink and UM DCO for their tireless computing support.
Toyota Research Institute (“TRI”) provided funds to assist the authors with their research but this article solely reflects the opinions and conclusions of its authors and not TRI or any other Toyota entity.

%% file: supp_arch.tex
\begin{table*}[t]
\caption{\textbf{Model Architecture}. Detailed model architecture, broken down into sub-components.
Please note, some components have more complicated structure, so we define operations at the beginning and forward pass at the bottom. For instance, in the Residual Module we define the two branches, followed by the forward pass calling each branch. We use $H=24$, $D=192$, $N_h=3$ in accordance with standard ViT-Tiny.
We do not define ResNet Blocks below, as we use the standard implementation available publicly.}\label{tab:model_arch}
\centering
\begin{tabular}{lc}
\toprule
\multicolumn{2}{c}{Overview} \\
 Operation & Output Shape           \\ \midrule
 Input Image & $ 2 \times 3 \times 256 \times 256$ \\
 Encoder & $ 2 \times H \times H \times D $           \\
 ViT Layer (x5) & $2 \times H \times H \times D$           \\
 Essential Matrix Module & $2 \times N_h \times (D/N_h + 6) \times (D/N_h + 6)$           \\
MLP & $7$           \\
\bottomrule
\end{tabular}

\vspace{2mm}

\begin{tabular}{lc}
\toprule
\multicolumn{2}{c}{Encoder} \\
 Operation & Output Shape           \\ \midrule
 ResNet-18 Block 1 & $ 2 \times 56 \times 56 \times 64$ \\
 ResNet-18 Block 2 & $ 2 \times H \times 28 \times 128$ \\
 Residual Module & $ 2 \times H \times H \times D$ \\
\bottomrule
\end{tabular}

\vspace{2mm}
\begin{tabular}{lc}
\toprule
\multicolumn{2}{c}{MLP} \\
 Operation & Output Shape           \\ \midrule
 Linear \& ReLU (x2) & 512 \\
 Linear \& ReLU & 7 \\
\bottomrule
\end{tabular}

\vspace{2mm}
\begin{tabular}{lc}
\toprule
\multicolumn{2}{c}{Residual Module} \\
 Operation & Output Shape           \\ \midrule
\textit{Branch A} \\
 2D Conv k=3 s=1, BN, ReLU & $ 2 \times 28 \times 28 \times D$ \\
 2D Conv k=5 s=1, BN, ReLU & $ 2 \times H \times H \times D$ \\
 \\
\textit{Branch B} \\
 2D Conv k=5 s=1, BN & $ 2 \times H \times H \times D$ \\
 \\
\textit{Forward Pass} \\
ReLU(Branch A + Branch B) & $2 \times H \times H \times D$ \\
\bottomrule
\end{tabular}
\end{table*}

\begin{table*}[t]
\caption{\textbf{Model Architecture: Essential Matrix Module.} Details of essential matrix module and ViT layer.}
\centering
\vspace{2mm}
\begin{tabular}{lc}
\toprule
\multicolumn{2}{c}{ViT Layer} \\
 Operation & Output Shape           \\ \midrule
{\it Attn} \\
Q,K,V = Linear & $ 2 \times N_h \times (H \times H) \times (D/N_h)$ \\
A = Softmax(Q @ K.T, dim=-1)     \quad \quad \quad \quad \quad                 \quad & $ 2 \times N_h \times (H \times H) \times (H \times H)$ \\
(A @ V).T & $ 2 \times H \times H \times D$ \\
Linear & $ 2 \times H \times H \times D$ \\
\\
\textit{MLP} \\
Linear \& GeLU & $ 2 \times H \times H \times (D \times 4)$ \\
Linear & $ 2 \times H \times H \times D$ \\
\\
\textit{Forward Pass} \\
 Attn(LayerNorm) + Residual & $ 2 \times H \times H \times D$ \\
 MLP(LayerNorm) + Residual & $ 2 \times H \times H \times D$ \\
\bottomrule
\end{tabular}

\vspace{2mm}
\begin{tabular}{lc}
\toprule
\multicolumn{2}{c}{Essential Matrix Module} \\
 Operation & Output Shape           \\ \midrule
\textit{MLP} \\
Linear \& GeLU & $ 2 \times H \times H \times (D \times 4)$ \\
Linear & $ 2 \times H \times H \times D$ \\
\\
\textit{Forward Pass} \\
 Essential Matrix Cross-Attn(LayerNorm) + Residual & $2 \times N_h \times (D/N_h + 6) \times (D/N_h + 6)$ \\
 MLP(LayerNorm) + Residual & $2 \times N_h \times (D/N_h + 6) \times (D/N_h + 6)$ \\
\bottomrule
\end{tabular}

\vspace{2mm}
\begin{tabular}{lc}
\toprule
\multicolumn{2}{c}{Essential Matrix Cross-Attn} \\
 Operation & Output Shape           \\ \midrule
\textit{Attn Branch} \\
A = Softmax(Q @ K.T, dim=-1) $\times$ Softmax(Q @ K.T, dim=-2) & $ N_h \times (H \times H) \times (H \times H)$ \\
(V.T @ A @ V).T & $N_h \times (D/N_h + 6) \times (D/N_h + 6)$ \\
Linear & $N_h \times (D/N_h + 6) \times (D/N_h + 6)$ \\
\\
\textit{Forward Pass} \\
$Q_1,K_1,V_1$ = Linear(Input$[0]$) & $ N_h \times (H \times H) \times (D/N_h + 6)$ \\
$Q_2,K_2,V_2$ = Linear(Input$[1]$) & $ N_h \times (H \times H) \times (D/N_h + 6)$ \\
$V_1[...,-6:], V_2[...,-6:]$ = Pos Encoding & $ N_h \times (H \times H) \times (D/N_h + 6)$ \\
Concat(Attn Branch($Q_1, K_2, V_2$), Attn Branch($Q_2, K_1, V_1$)) & $2 \times N_h \times (D/N_h + 6) \times (D/N_h + 6)$ \\
\bottomrule
\end{tabular}

\end{table*}

Our architecture is outlined below and detailed in the following two-page Table \ref{tab:model_arch}. 

\parnobf{Backbone} Our backbone consists of a vanilla encoder of Residual Modules and one Block, followed by a vanilla ViT. The model ends with a three-layer MLP.

\parnobf{Essential Matrix Module} Recall from the paper, the Essential Matrix Module is closely built off a standard Cross-Attention Block.
The changes we make are carefully chosen to study their contribution to performance.
Changes can be seen in the ``Essential Matrix Cross-Attn'' block of the architecture.
First, the final dimension of query, key and values is increased by a size of 6; positional encodings fill these 6 new spaces for value matrices.
Second, Softmax is computed over both the last and second to last dimension of affinities, and multiplied elementwise.
Finally, Bi-Directional attention is computed, $V.T @ A @ V$.
Note this results in output of different shape than standard Cross-Attention.

\parnobf{Baselines} Baselines use generally the same components as our entire architecture, with very minor changes. This helps us study our proposed contributions.

\parnoit{CNN Pose Regressor} CNN Pose Regressor follows our model architecture, with the exception it only uses encoder, and one pooling layer, before the MLP.
The pooling layer consists of two 1x1 Convs with Batchnorm (and ReLU before the second Conv). 
Feature size is reduced to 96, then 43; resulting in MLP input of $24768=43*24*24$.
We do this so the input to the MLP is comparable to our method ($29400=2*3*(64+6)*(64+6)$), and so this method is less prone to poor overfitting.
Early experiments with bigger input size to MLP hurt performance.
The CNN and MLP are otherwise identical to ours.

\parnoit{+ViT} Starting from the CNN Pose Regressor, we simply add the ViT Layers back from our architecture.
This is followed by a vanilla Cross-Attention block. This uses the same pooling layer as the CNN.
The Cross Attention-Block can be distinguished from our Essential Matrix Module by changing bidirectional attention, dual softmax and positional encoding.
Looking in the table, the differences can be found in Essential Matrix Cross-Attn.
First, affinities are calculated using the standard $(A @ V).T$.
Second, Softmax is not applied over $dim=-2$.
Third, values do not get positional encodings.

\parnoit{+Bilinear Attention} This method uses our architecture as-is, with the exception of dual softmax and positional encoding.
Looking in the table, the differences can be found in Essential Matrix Cross-Attn.
First, Softmax is not applied over $dim=-2$.
Second, values will not get positional encodings.

\parnoit{+Dual Softmax} This method uses our architecture as-is, with the exception of positional encoding.
Looking in the table, the difference can be found in Essential Matrix Cross-Attn -- values will no longer get positional encodings.

%% file: supp_dataset.tex
\parnobf{Matterport3D} Matterport3D is a collection of scanned indoor scenes.
We use the image pairs Jin \textit{et al.} collected using the Habitat simulator.
The number of image pairs in the train, val and test set are 31932, 4707, and 7996, respectively.
Images are originally 480x640, but are downsampled to 256x256 for models.
Images are collected using a camera at random height of 1.5-1.6m, with a downward tilt of 11 degrees to simulate
human perspective. 
Candidate pairs are randomly sampled cameras within each room.
Next, Jin \textit{et al.} detect planes, and select pairs such that at least 3 planes are shared between images, and at least 3 planes are unique to each image. 
The average rotation is 53 degrees, translation 2.3m, and overlap 21\%.

\parnobf{InteriorNet} InteriorNet consists of 10,050 indoor panoramic views across 112 synthetic houses.
82 houses are used for training and 30 are used for testing. 
We sample paired images from the panoramas using the same procedure as Cai \textit{et al.} 
We also follow their image selection process, which samples images over a uniform distribution of angles (yaw in [-180, 180]; pitch in [-30,30]) within panoramas.
Their process samples 100 images per panorama, filters images too close to walls, and does not apply roll; arguing this does not affect performance. 
Images are 256x256 and have 90$^{\circ}$ FOV.
For full details see the original paper.

For the InteriorNet-T dataset, pairs are selected from different panoramas, resulting in translation;
for InteriorNet, pairs are selected from the same panorama, resulting in no translation.
Translations in InteriorNet-T are selected to be less than 3m.
The full set of extracted image pairs on InteriorNet is ~250K (~610K for InteriorNet-T). 
Note these train set sizes are smaller than reported in Cai \textit{et al.} (roughly ~1M and ~700k) but were supplied by the authors directly and via their repo; these sets replicate their reported paper results.
Both have a test set of 1K pairs.
We consider only overlapping pairs, making the train set smaller: InteriorNet: 150K, InteriorNet-T: 350K.
The test set sizes are also reduced after filtering for overlap. Pairs are further broken down into large or small overlap, thresholded using rotation of 45 degrees.
InteriorNet test set: 695 pairs (403 large overlap, 292 small overlap).
InteriorNet-T test set: 659 pairs (335 large overlap, 324 small overlap).

\parnobf{StreetLearn} StreetLearn consists of 143K panoramic outdoor views in New York City and Pittsburgh;
we follow the setup of Cai \textit{et al.} which uses 56K panoramas from Manhattan, with 1K randomly allocated for testing.
We also follow their image selection process, which samples images over a uniform distribution of angles (yaw in [-180, 180]; pitch in [-45,45]) within panoramas.
Their process samples 100 images per panorama, and does not apply roll; arguing this does not affect performance. 
Images are 256x256 and have 90$^{\circ}$ FOV.
For full details see the original paper.
For the StreetLearn-T dataset, pairs are selected from different panoramas, resulting in translation;
for StreetLearn, pairs are selected from the same panorama, resulting in no translation.
Translations in StreetLearn-T are selected to be less than 10m.
The full set of extracted image pairs on StreetLearn is ~1.1M (670K for StreetLearn-T). Both have a test set of 1K pairs.
We consider only overlapping pairs, making the train set smaller: StreetLearn: ~460K, StreetLearn-T: ~260K.
The test set sizes are also reduced after filtering for overlap. Pairs are further broken down into large or small overlap, thresholded using rotation of 45 degrees.
StreetLearn test set: 471 pairs (170 large overlap, 301 small overlap).
StreetLearn-T test set: 469 pairs (152 large overlap, 317 small overlap).

StreetLearn panoramas were captured based on Google Street View, and therefore contain real people. 
The authors of the dataset blurred all faces and license plates, and manually reviewed images for privacy.
The dataset is distributed only upon request.
If individuals request for a panorama to be taken down or blurred, the dataset is updated to reflect the request.

%% file: supp_additional.tex
\begin{figure}[t]
	\centering
			{\includegraphics[width=0.475\linewidth]{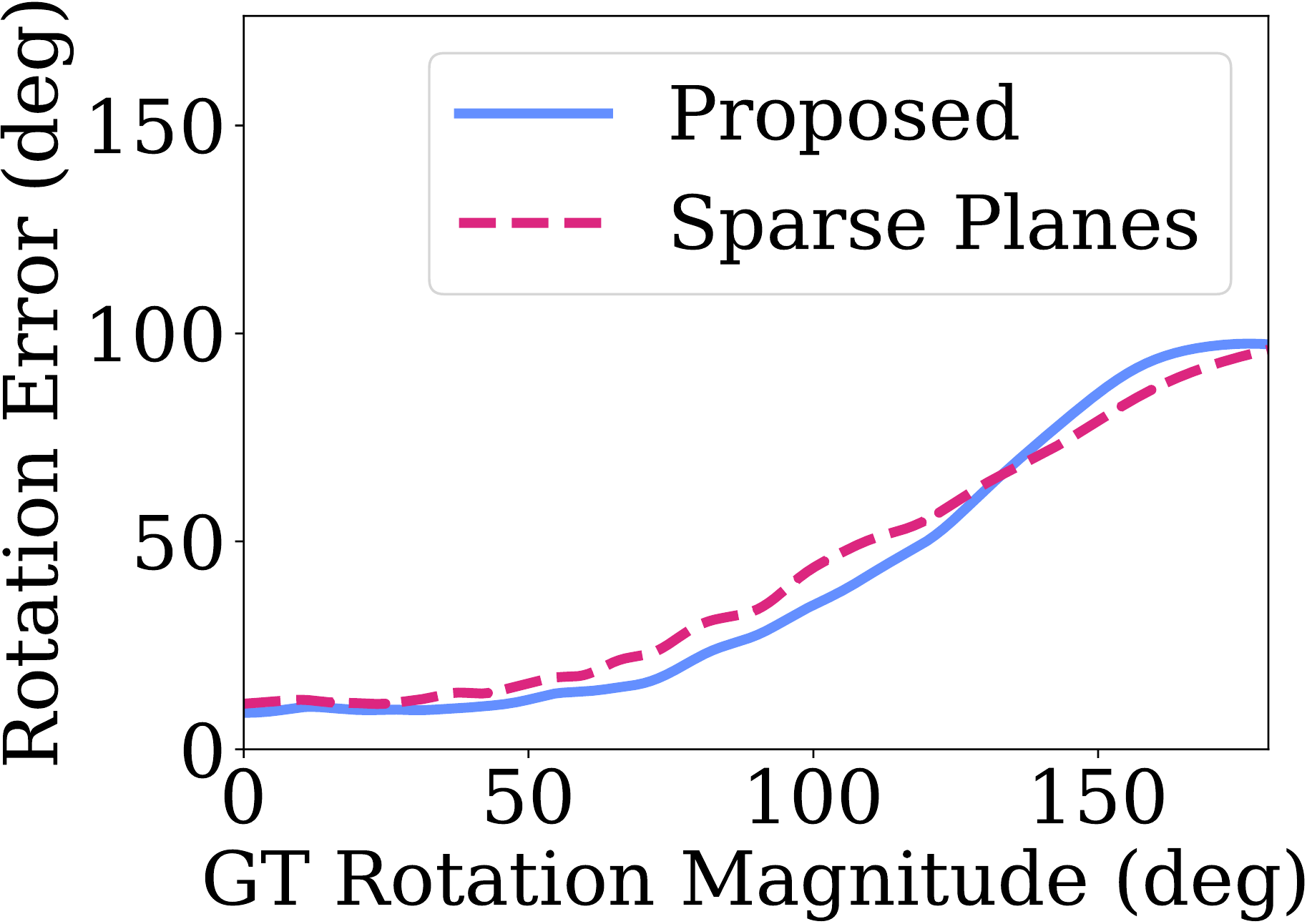}}
	\caption{\textbf{Mean Error as a Function of Magnitude: Rotation.}}
	\label{fig:er_rot}
\end{figure}
\begin{figure}[t]
	\centering
			{\includegraphics[width=0.475\linewidth]{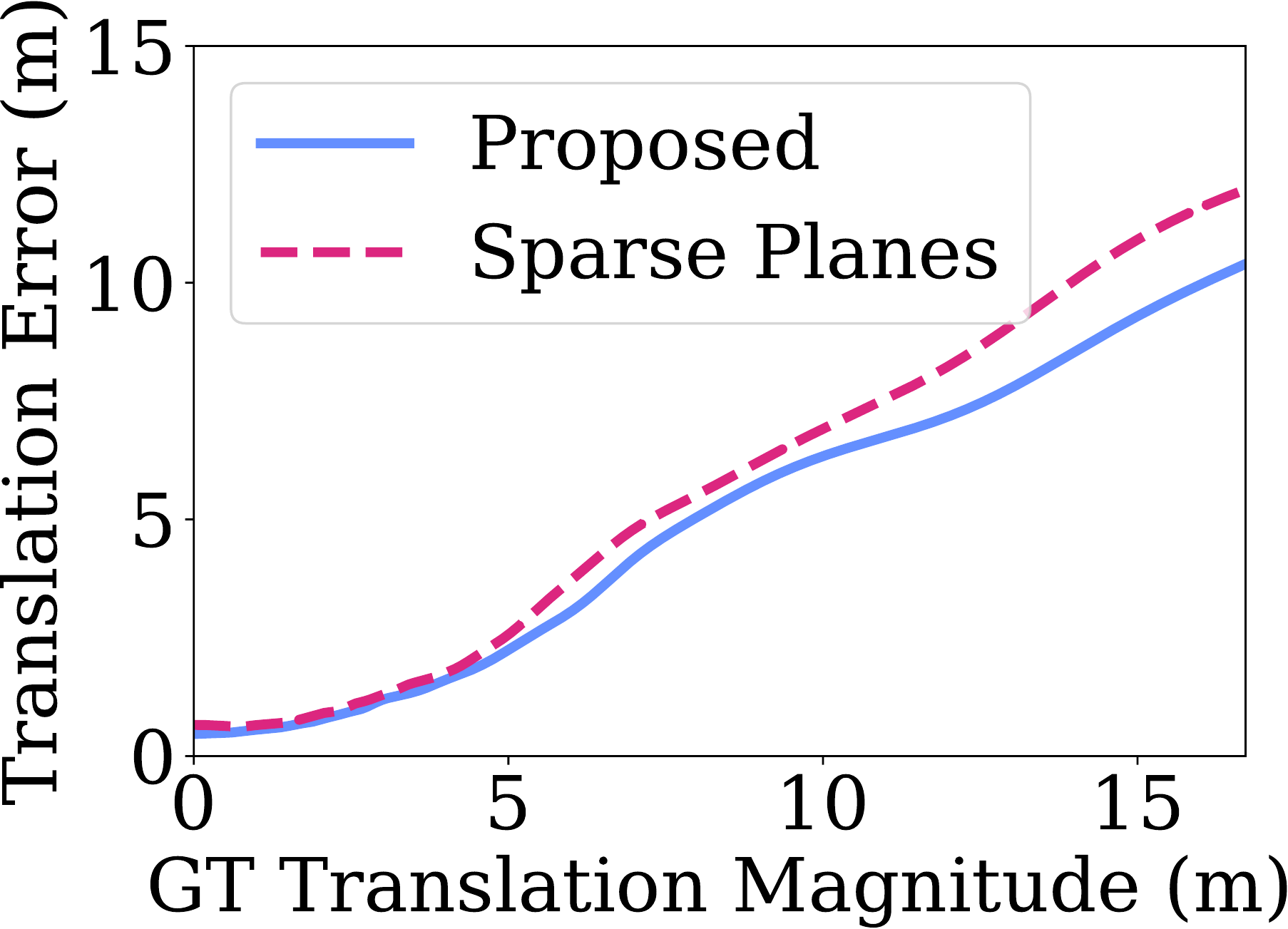}}
	\caption{\textbf{Mean Error as a Function of Magnitude: Translation.}}
	\label{fig:er_tran}
\end{figure}

\parnobf{Error against GT} In Figures \ref{fig:er_rot} and \ref{fig:er_tran}, we report error vs. magnitude for rotation and translation, respectively; for our method and baseline Sparse Planes. 
Lines are fit by applying adaptive kernel density estimation.
For both rotation and translation, both methods show similar general trends, increasing error as magnitude increases.
In the case of very small rotations (i.e. $<30^{\circ}$) or translations (i.e. $<3m$), Sparse Planes is quite competitive with the proposed method. 
However, beyond this magnitude, our method tends to be more robust, outperforming until the very most extreme rotations. 
This is consistent with the qualitative results and CDFs plotted in the paper.

\parnobf{Generalization Across Datasets} A good measure of inductive bias is to evaluate methods when trained on one dataset and tested without fine-tuning on another.
We evaluate our method against the most competitive baselines across datasets in Table~\ref{tab:gen}.
We tend to generalize better, especially in average error, though indoor $\xleftrightarrow{}$ outdoor does cause a large drop in performance. 

\begin{table}
\caption{\textbf{Generalization Across Datasets}, training Cai \textit{et al.} and Ours on the opposite of IN-T and SL-T. SuperGlue is trained on ScanNet; the only public version. In format ``Large / Small'' Overlap.}
\label{tab:gen}
\resizebox{\ifdim\width>\linewidth \linewidth \else \width \fi}{!}{
\begin{tabular}{lcccccc}
\toprule 
 & \multicolumn{3}{c}{InteriorNet-T} & \multicolumn{3}{c}{StreetLearn-T} \\
Method & Avg ($^{\circ}$ $\downarrow$) & Med. ($^{\circ}$ $\downarrow$) & 10 (\% $\uparrow$) & Avg ($^{\circ}$ $\downarrow$) & Med. ($^{\circ}$ $\downarrow$) & 10 (\% $\uparrow$) \\ 

\midrule
SuperGlue & 35.9 / 85.4 & 35.9 / 81.6 & 5.4 / 0.0 & 48.9 / 85.5 & 42.9 / 82.4 & 2.0 / 0.0 \\
Cai \textit{et al.} & 31.2 / 67.4 & \textbf{9.3} / 84.4 &\textbf{52.0} / \textbf{23.8} & 43.0 / 71.4 & 21.8 / 72.5 & 32.4 / 9.9  \\ 
Ours & \textbf{18.7} / \textbf{58.6} & 11.1 / \textbf{66.4} & 45.1 / 10.5 & \textbf{28.6} / \textbf{40.7} & \textbf{14.3} / \textbf{24.0} & \textbf{39.5} / \textbf{27.8} \\ 
\bottomrule
\end{tabular}
}
\end{table}

\begin{figure*}
\includegraphics[width=\linewidth]{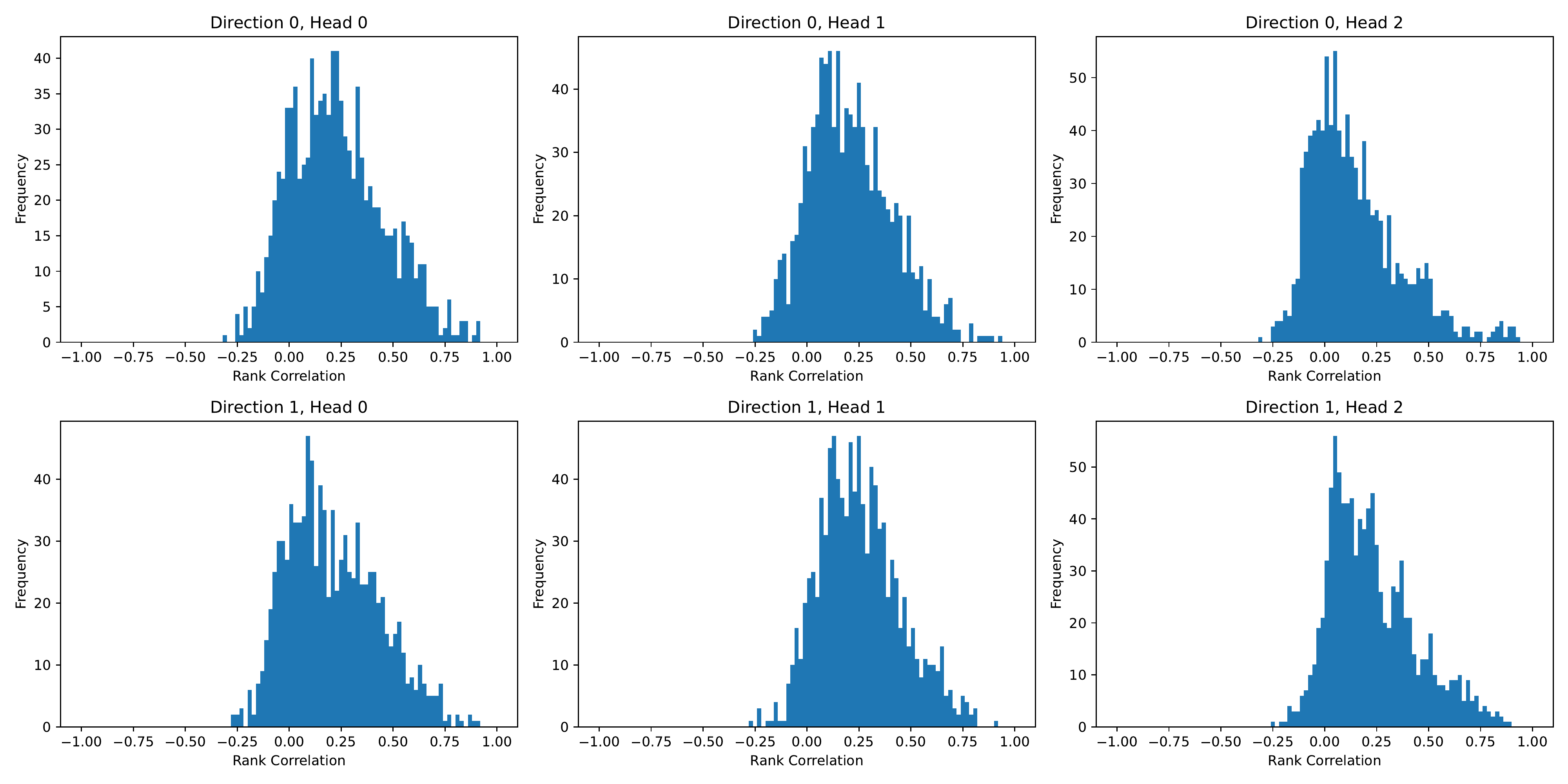} 
\caption{Distribution of rank correlations between ground-truth $\UB^\top \UB$ 
and the $\UB^\top \UB$ as computed by the transformer. We re-arrange the $6\times6$ bottom
sub-matrix to be $9\times9$ and then compute the Spearman rank correlation per-image (which ranges
from -1 to 1). We plot histograms of these rank correlations for each direction (i.e., image 1 or image 2) and
each of the three transformer heads. Even though the matrices computed by the transformers are not used in the
same way as in the 8-point algorithm, we find substantial rank correlation between the two.}
\label{fig:utu}
\end{figure*}

\parnobfnoper{Is ViT Backbone Needed?} Our proposed method appends an Essential Matrix Module layer to the end of 5 ViT layers.
In Table~\ref{tab:bb}, we study the effect of replacing the 5 ViT layers with a comparable CNN-based backbone, a modified PWC-Net~\cite{sun2018pwc}.
PWC-Net is a good choice as it predicts optical flow, a task closely related to correspondence estimation and therefore useful for downstream pose estimation;
TartanVO~\cite{wang2021tartanvo} uses this extractor before predicting relative pose.
We use PWC-Net out of the box except for modifications to feature and stride size so that it takes extracted features at the same resolution (24x24) and feature size (192) as the ViT, and produces output features at this same resolution and feature size.
The CNN-based backbone performs similarly to the ViT backbone -- better on StreetLearn-T but worse on InteriorNet-T.
This indicates the Essential Matrix Module may be flexible for use with more generic models.
Recall from Tables 3, 5 and 6 in the main paper, the ViT backbone is not the primary reason for our success -- the Essential Matrix Module is a helpful inductive bias, particularly in the case of limited data.

\begin{table}
    \caption{\textbf{Model Backbone Ablations.} A ViT backbone is not necessary for competitive performance; a CNN replacement performs similarly if we keep the Essential Matrix Module.}
\label{tab:bb}
\resizebox{\ifdim\width>\linewidth \linewidth \else \width \fi}{!}{
\begin{tabular}{l l c c c c c c} \toprule
  & & \multicolumn{3}{c}{InteriorNet-T} & \multicolumn{3}{c}{StreetLearn-T} \\
Overlap & Method & Avg ($^{\circ}$ $\downarrow$) & Med. ($^{\circ}$ $\downarrow$) & 10 (\% $\uparrow$) & Avg ($^{\circ}$ $\downarrow$) & Med. ($^{\circ}$ $\downarrow$) & 10 (\% $\uparrow$) \\
\midrule
\multirow{2}{*}{Large} & CNN & 5.36 & 3.62 & 94.03 & \textbf{3.74} & \textbf{2.29} & \textbf{94.74} \\
& ViT (Ours) & \textbf{2.90} & \textbf{1.83} & \textbf{97.91} & 4.08 & 2.43 & 90.13 \\
\midrule
\multirow{2}{*}{Small} & CNN & 8.31 & 4.58 & 86.42 & \textbf{7.30} & \textbf{3.03} & \textbf{90.54} \\
& ViT (Ours) & \textbf{4.48} & \textbf{2.38} & \textbf{96.30} & 9.19 & 3.25 & 87.70 \\
\bottomrule
\end{tabular}
}
\end{table}

\parnobfnoper{Can one Construct the Essential Matrix Module from Conv Kernels?}
One can construct the essential matrix from $\PhiB^\top \AB \PhiB$, which is similar to the attention operation performed in ViTs but not to any operation in Conv layers. 

\parnobfnoper{Can Essential Matrix Module Output be used for the 8-Point Algorithm?} 
No, it is an inductive bias used for pose prediction. 
An analogy is convolution layers, which have the capacity to detect edges but which do not necessarily just detect edges when learned.
Likewise, the EM Module has the capacity to represent information like $\UB^\top \UB$, from which one could compute the Essential matrix. 
Our results show that this inductive bias helps learned pose estimation.
We also note that some of the EM Module's output (the upper $D \times D$ submatrix) are image features that cannot be used in the 8-Point algorithm.

\parnobfnoper{How closely does EMM's $\sim\UB^\top \UB$ match true $\UB^\top \UB$?} We save EMM $\sim\UB^\top \UB$ output for the Matterport test set and compare to true $\UB^\top \UB$, computed by projecting 100k points in a 10x10x10 box into both images using ground truth pose. Note the EMM computes cross-attention in both img1 $\to$ img2 and img2 $\to$ img1, and has attention 3 heads for each direction. We therefore have 6 $\sim\UB^\top \UB$ to compare to the ground truth. We measure rank correlation between entries of $\sim\UB^\top \UB$ and $\UB^\top \UB$.

As we see in Figure~\ref{fig:utu} (total correlation values), there is a nontrivial rank correlation between EMM $\sim\UB^\top \UB$ and the true $\UB^\top \UB$. 
This correlation varies some on a per-head and per-direction basis, which is not surprising given directions and heads are concatenated before pose regression, and may be used for differing purposes and extents.
Beyond concatenating heads, one important reason correlation is not higher is that $\sim\UB^\top \UB$ is followed by linear projection and normalization layers before pose regression. Since there is no constraint on $\sim\UB^\top \UB$ to match the true $\UB^\top \UB$, the end-to-end system is instead encouraged to learn pose optimally, with this structure used as an inductive bias. Nevertheless, the positive correlation between $\sim\UB^\top \UB$ and $\UB^\top \UB$ is consistent with our expectations the EMM is in fact computing similar positional structure to the eight point algorithm.

\begin{figure*}
\centering
\begin{tabular}{c}
\includegraphics[width=0.8\linewidth]{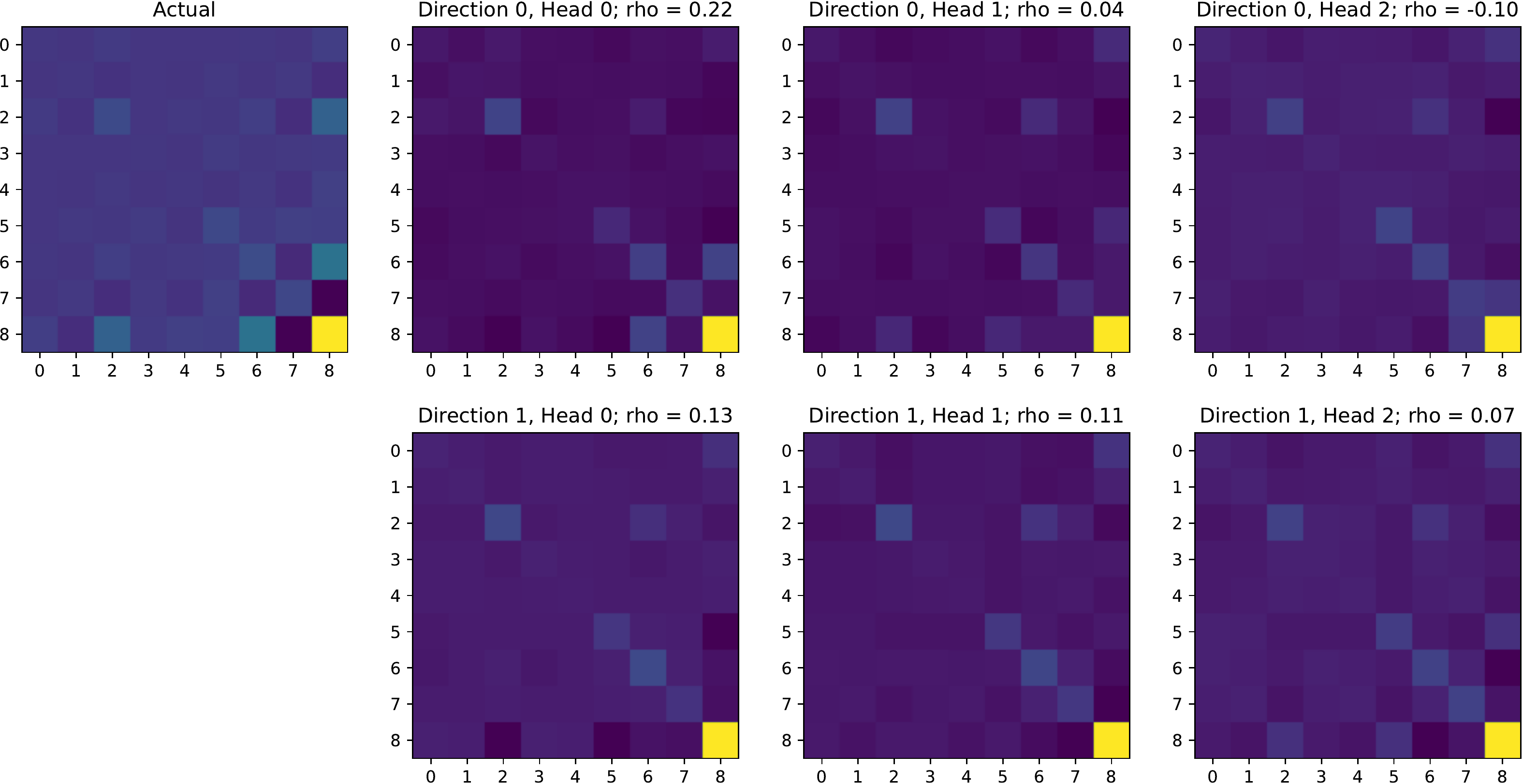} \\
\includegraphics[width=0.8\linewidth]{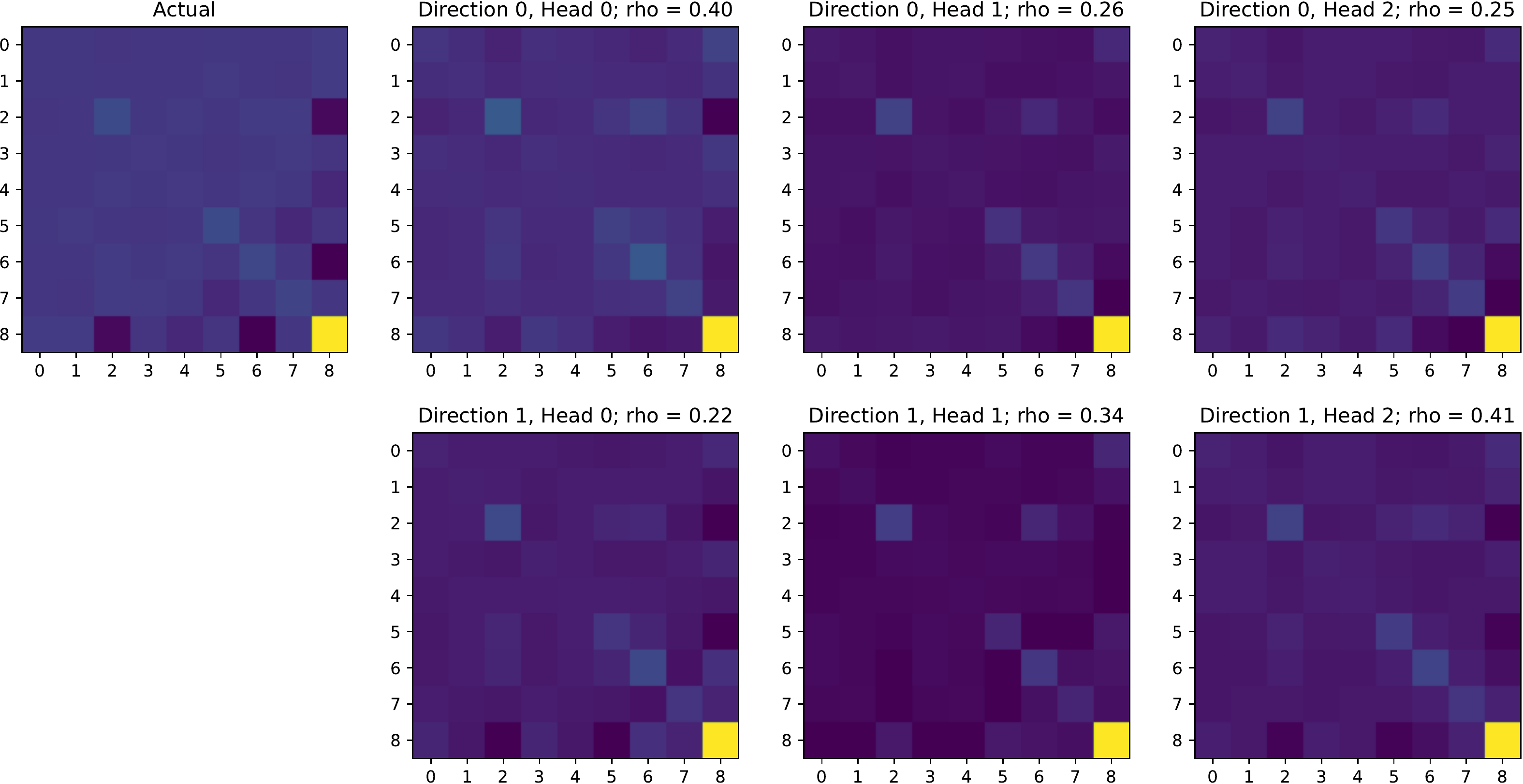} \\
\includegraphics[width=0.8\linewidth]{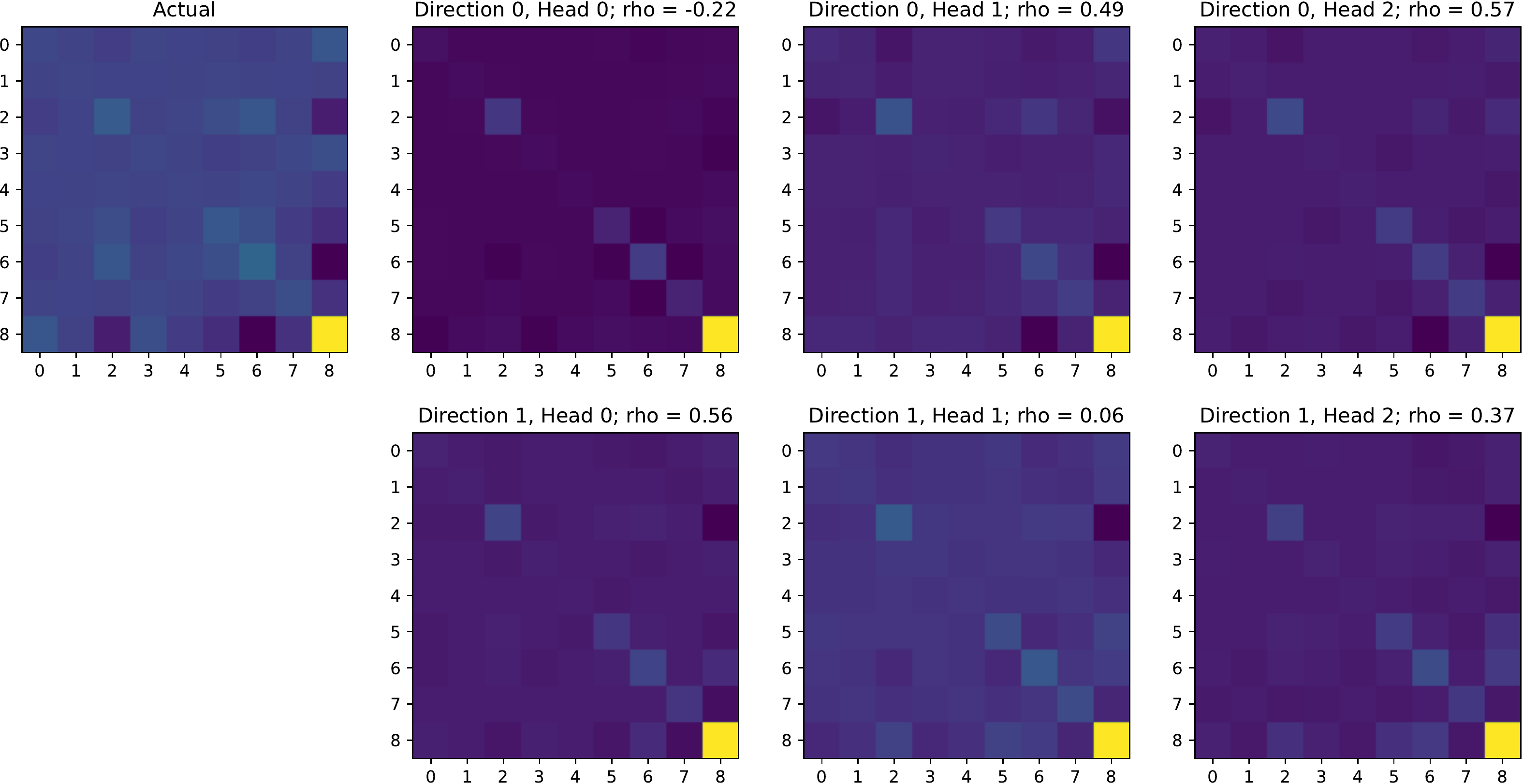} \\
\label{fig:egs}
\end{tabular}
\caption{Selected examples of the actual and transformer $\UB^\top \UB$ matrices for three examples. Top left: actual $\UB^\top \UB$. Each row shows a different direction (i.e., image 1 or image 2) and each column
shows a different transformer head (i.e., one of the three heads). The transformer matrix is actually $6 \times 6$, but we re-arrange it to be the $9 \times 9$ $\UB^\top \UB$ matrix.}
\end{figure*}

We visualize $\sim\UB^\top \UB$ output of individual image pairs in Figure~\ref{fig:egs}. 
Here we see clear examples of the varying, but mostly positive and meaningful, correlation between actual $\sim\UB^\top \UB$ and output from different heads and directions.
In addition, comparison of individual outputs in the $9x9$ $\sim\UB^\top \UB$ tends to show similar cells with large activation across heads, regardless of actual correlation.
We emphasize we should not expect $\sim\UB^\top \UB$ to precisely match $\UB^\top \UB$.

\parnobfnoper{Why not use 8-Point Coefficients as $\PhiB$?} 
Using 8-point coefficients as $\PhiB$ (Eqn.~\ref{eqn:ten}) actually leads to the same attention output entries as ours (Eqn.~\ref{eqn:eight}), only ours has the advantage of being more compact, as it doesn’t duplicate entries.
This is detailed in Sec.~\ref{sec:matrixverify}.

\parnobfnoper{How is Scale Ambiguity Handled?} The model is trained end-to-end using translation loss with scale, leaving the network free to learn to overcome the scale ambiguity via recognition. 
We believe this is due to a mix of a distribution over likely relative poses inside scenes as well as familiar objects. 
Essential Matrix Module output mixes learned appearance features $\VB$ (which can encode familiar objects), positional encodings $\PhiB$, and their interaction. 
Since $\PhiB$ uses intrinsics (L476), features can model varying cameras. 

\begin{figure*}[t]
	\centering
			{\includegraphics[width=\linewidth]{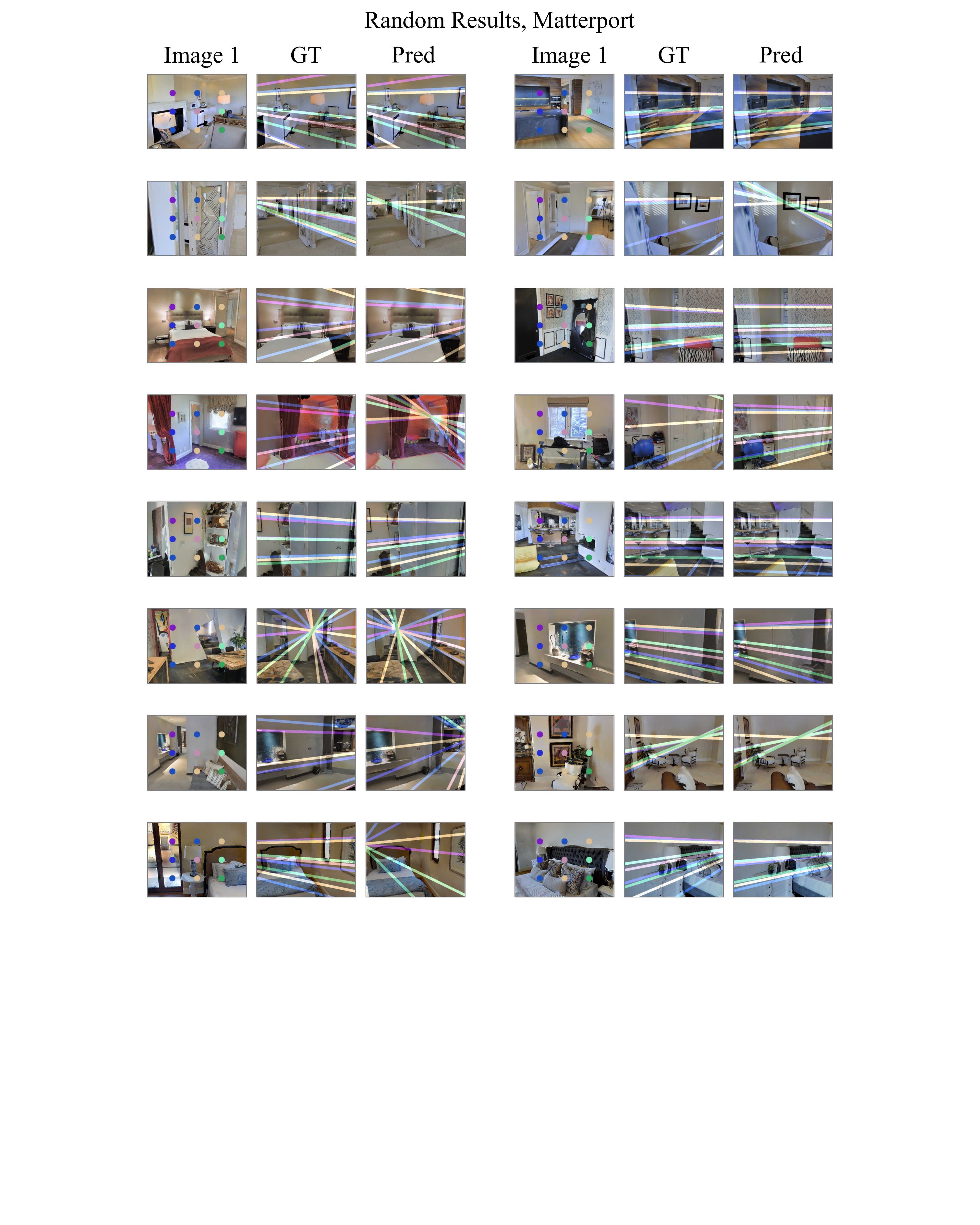}}
	\caption{\textbf{Random Results on Matterport: Epipolar Lines.}}
	\label{fig:random_epipolar_matterport}
\end{figure*}

\begin{figure*}[t]
	\centering
			{\includegraphics[width=\linewidth]{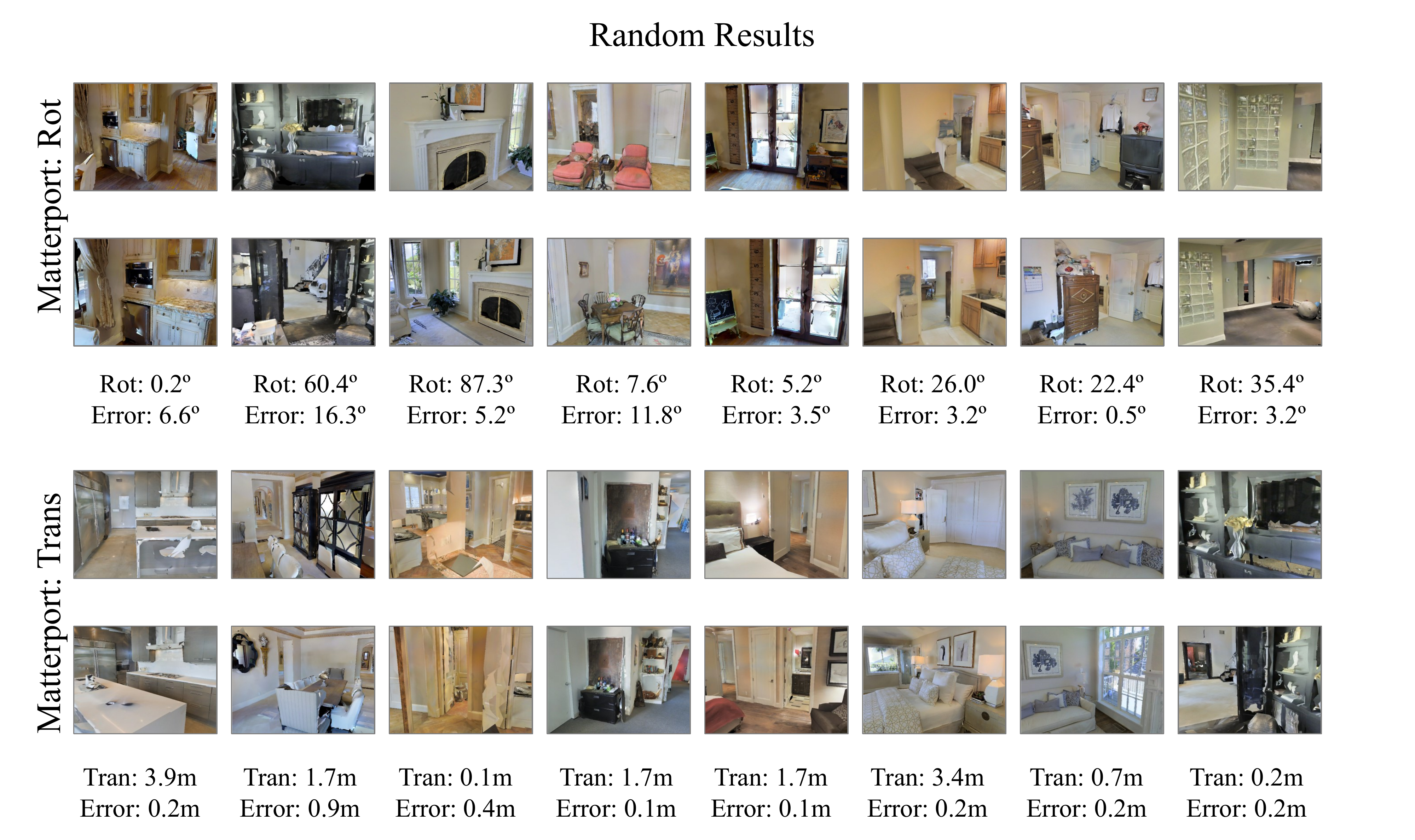}}
	\caption{\textbf{Random Results on Matterport.}}
	\label{fig:random_matterport}
\end{figure*}

\begin{figure*}[t]
	\centering
			{\includegraphics[width=\linewidth]{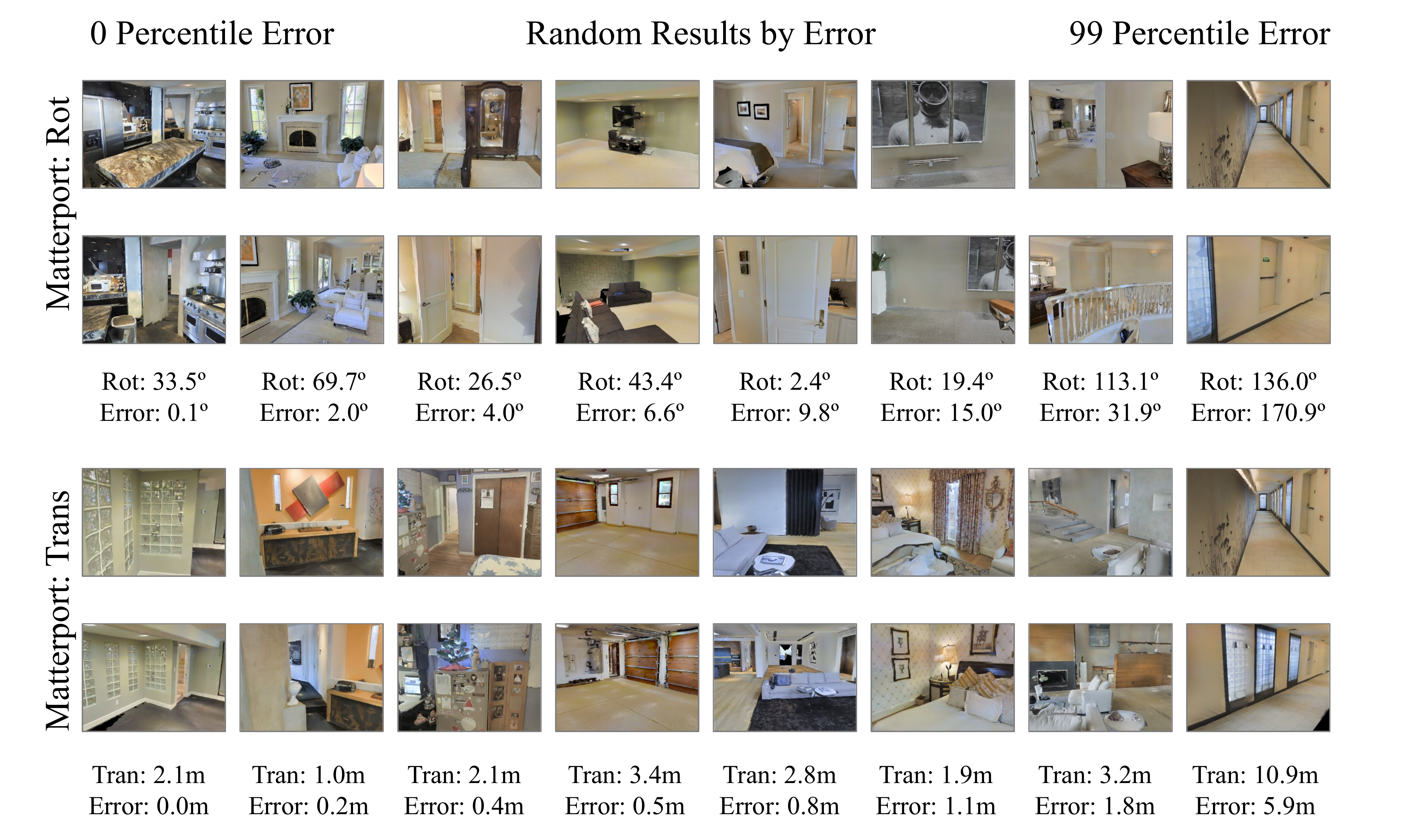}}
	\caption{\textbf{Results by error on Matterport.}}
	\label{fig:error_matterport}
\end{figure*}

\parnobf{Additional Qualitative Results: Matterport} In Figure \ref{fig:random_epipolar_matterport}, we see the proposed method generates Epipolar lines generally similar to true view changes.
This is consistent with random rotation and translation errors in Figure \ref{fig:random_matterport}, which are generally small.
Not surprisingly, random results in this challenging setting are also sometimes poor, e.g. the top left Epipolar example, or the top left rotation error example. 
Sorting by error in Figure \ref{fig:error_matterport}, we see the general trend of increased view change being associated with increased error, which is consistent with our Error against GT study.

\begin{figure*}[t]
	\centering
			{\includegraphics[width=\linewidth]{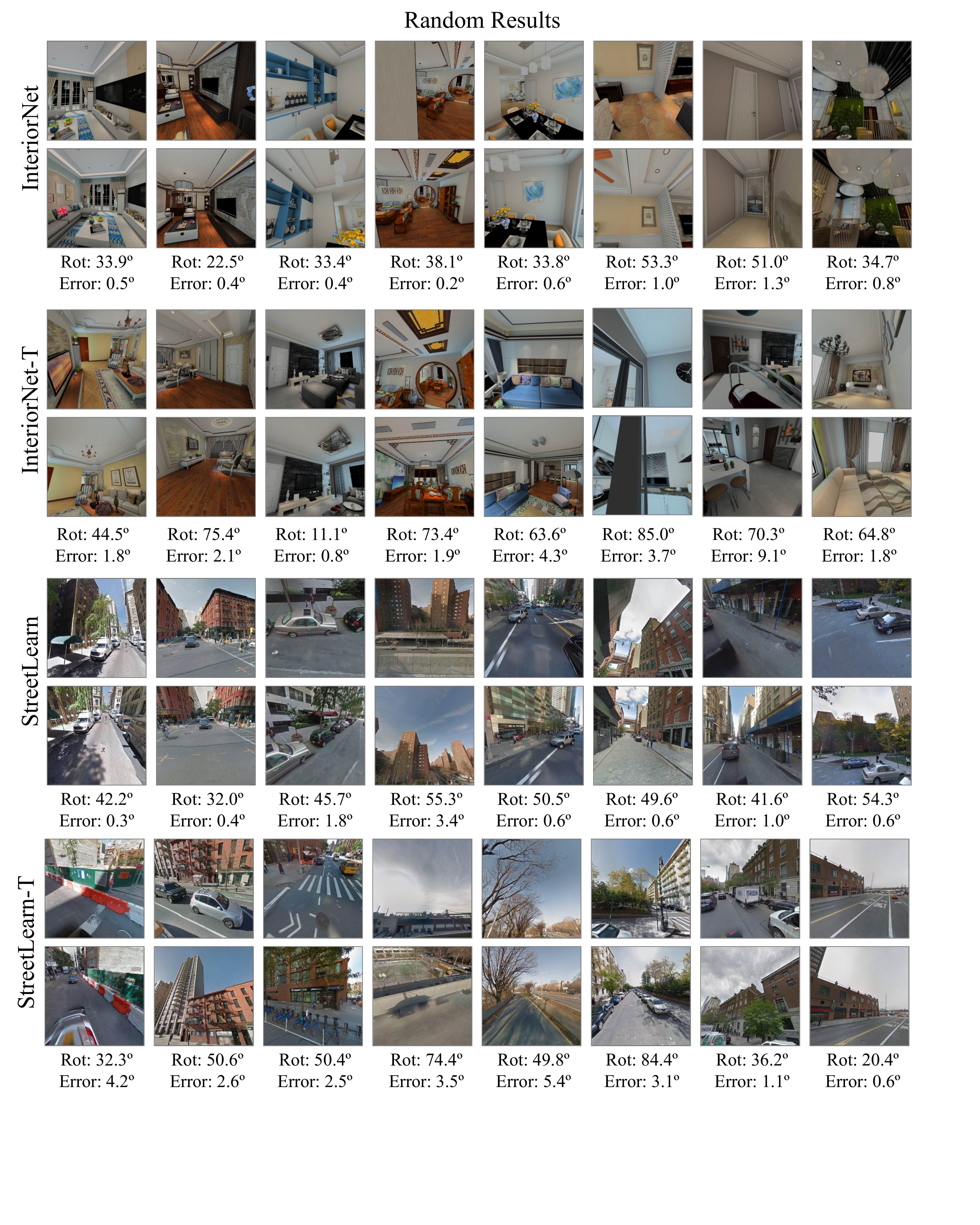}}
	\caption{\textbf{Random Results on InteriorNet and StreetLearn.}}
	\label{fig:random_interiornet_streetlearn}
\end{figure*}

\begin{figure*}[t]
	\centering
			{\includegraphics[width=\linewidth]{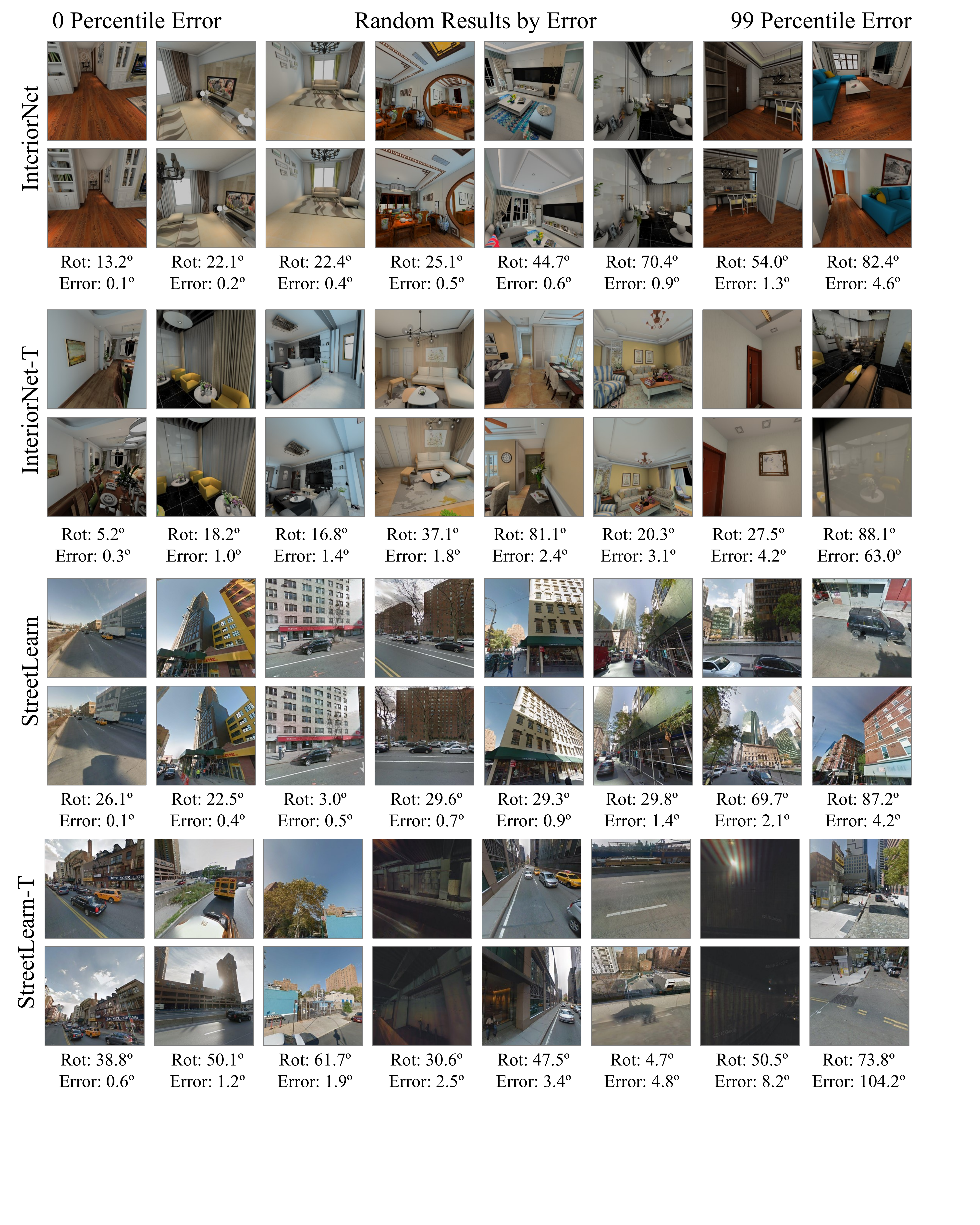}}
	\caption{\textbf{Results by error on InteriorNet and StreetLearn.}}
	\label{fig:error_interiornet_streetlearn}
\end{figure*}

\parnobf{Additional Qualitative Results: InteriorNet and StreetLearn} In Figure \ref{fig:random_interiornet_streetlearn}, we see random rotation and translation errors, which are generally quite small.
Random results in these datasets are not often poor, but have some weaker examples -- e.g. the bottom row of StreetLearn-T has most errors above 1$^{\circ}$.
Sorting by error in Figure \ref{fig:error_interiornet_streetlearn}, we again see the trend of increased view change being associated with increased error.
Yet, results are often quite good even in very high rotations (e.g. InteriorNet all examples).

%% file: supp_deriv.tex
Our goal is to show that the unique entries of $\UB^\top \UB$ that is used in the Eight Point Algorithm can be computed 
as $\PhiB^\top \AB \PhiB$ for an attention matrix $\AB \in \{0,1\}^{P \times P}$ and 
$\PhiB \in \mathbb{R}^{P \times 6}$ as defined in the main paper.

\parnobf{Setup}
Given $N$ correspondences, the eight-point algorithm constructs a 
matrix $\UB \in \mathbb{R}^{N \times 9}$
row-wise via the Kronecker products of the homogeneous coordinates of the correspondences involved.
Define $\xB_i = [u_i, v_i, 1]$ and $\xB' = [u'_i, v'_i, 1]$. Then 
the $i$th row of $\UB$ is 
\begin{equation}
\UB_{i,:} = \left[ 
\begin{array}{ccccccccc}
u_i u'_i &
u_i v'_i &
u_i &
v_i u'_i &
v_i v'_i &
v_i &
u'_i &
v'_i &
1
\end{array}
\right]
\end{equation}
or more compactly,
\begin{equation}
\UB_{i,:} = (\xB_i \otimes \xB'_i)^\top.
\end{equation}
Note that when estimating the Essential matrix, one uses $\xB_i \equiv \KB^{-1} [u_i, v_i, 1]^\top$
and $\xB'_i \equiv \KB^{-1} [u'_i, v'_i, 1]^\top$. Since these coordinates can be rescaled by any 
arbitrary non-zero scalar, we assume that the last coordinate is $1$. 

\parnobf{Usual approach} Given correct correspondences, the eigenvector of $\UB^\top \UB \in
\mathbb{R}^{9 \times 9}$ that corresponds to the smallest eigenvector is the
Essential or Fundamental matrix. Usually, the matrix is not rank-deficient, and so one
reshapes the eigenvector and then performs rank-reduction.

\parnobf{Alternate approach}
We will now show that the unique entries of $\UB^\top \UB$ can be computed in an alternate
fashion using a setup that is amenable to computation via a transformer. We'll start
with the following basic substitutions and cleaning up:
\begin{equation}
\UB^\top \UB = \sum_{i=1}^N \UB_{i,:}^\top \UB_{i,:} = \sum_{i=1}^N (\xB_i \otimes \xB'_i) (\xB_i \otimes \xB'_i)^\top.
\end{equation}
We'll first rewrite the interior of the sum, and then the sum itself.

\parnoit{Rewriting $\UB_{i,:}^\top \UB_{i,:}$ with a basis expansion}
We'll tackle the interior of the sum first. While 
$\UB_{i,:}^\top \UB_{i,:} = (\xB_i \otimes \xB'_i) (\xB_i \otimes \xB'_i)^\top \in \mathbb{R}^{9 \times 9}$ and thus has 81 entries,
there are only 36 unique entries. The smaller number of entries can be seen mechanically via direct expansion
(see \S\ref{sec:matrixverify} to see this). It can also be reasoned out by 
distributing transposes and using the mixed product property to rewrite it as
\begin{equation}
(\xB_i \otimes \xB'_i) (\xB_i \otimes \xB'_i)^\top =  (\xB_i \xB_i^\top) \otimes (\xB'_i {\xB'_i}^\top).
\end{equation}
Note that while $\xB_i \xB_i^\top$ has 9 entries, it only has 6 {\it unique} entries ($1$, $u$, $v$, $uv$, $u^2$, $v^2$).
Likewise, $\xB'_i {\xB'_i}^\top$ has 6 unique entries ($1$, $u'$, $v'$, $u'v'$, ${u'}^2$, ${v'}^2$). Therefore, their Kronecker product 
$(\xB_i \xB_i^\top) \otimes (\xB'_i {\xB'_i}^\top)$ has only 36 unique entries.

We can create a $6 \times 6$ matrix containing the unique entries of $\UB_{i,:}^\top \UB_{i,:}$ by applying
a basis expansion to the coordinates. Let us define $\phi([u,v,1]) = [1, u, v, uv, u^2, v^2]$. Then
the unique entries of $\UB_{i,:}^\top \UB_{i,:} \in \mathbb{R}^{9 \times 9}$ can be written as 
$\phi(\xB_i) \phi(\xB'_i)^\top \in \mathbb{R}^{6 \times 6}$. This means that the unique
entries of $\UB^\top \UB$ are given by 
\begin{equation}
\label{eqn:uniqueentires}
\sum_{i=1}^N \phi(\xB_i) \phi(\xB'_i)^\top.
\end{equation}
As an additional benefit, this factorization separates the terms involving each image into separate components.

\parnoit{Making the sum implicit}
We next rewrite the sum implicitly by assuming each correspondence lies on a fixed grid. Given
a grid of $P$ patches in each image, we assume that $\pB_j$ is the jth patch's location. Then, 
rather than have $N$ correspondences, we can define the correspondences implicitly via an
indicator matrix $\AB \in \{0,1\}^{P \times P}$ such that $\AB_{j,k} = 1$ if and only if points $\pB_k$ and $\pB_j$ are in
correspondence and 0 otherwise. If each correspondence is on each patch, then we can rewrite
\begin{equation}
\label{eqn:implicitsum}
\sum_{i=1}^N \phi(\xB_i) \phi(\xB'_i)^\top = 
\sum_{j=1}^P \sum_{k=1}^P \phi(\pB_k) \AB_{j,k} \phi(\pB'_j)^\top.
\end{equation}
This can be
further simplified by gathering the basis expanded coordinates of the grid in a matrix 
$\PhiB \in \mathbb{R}^{P \times 6}$
such that 
$\PhiB_{j,:} = \phi(\pB_j)^\top$. Then $\phi(\pB_k) \AB_{j,k} \phi(\pB'_j)^\top = \PhiB_{k,:}^\top \AB_{j,k} \PhiB_{j,:}$,
and so Equation~\ref{eqn:implicitsum} can be rewritten as
\begin{equation}
\sum_{j=1}^P \sum_{k=1}^P \PhiB_{k,:}^\top \AB_{j,k} \PhiB_{j,:} = \PhiB^\top \AB \PhiB, \\
\end{equation}
and therefore the unique entries of $\UB^\top \UB$ can be compactly written as $\PhiB^\top \AB \PhiB$.

%% file: supp_limitations.tex
The $\PhiB^\top \AB \PhiB$ expression is exact when: (1) every correspondence can be represented as one of the $P$ patches;
and (2) the attention matrix $\AB$ produced by the ViT represents correspondence and is binary. Without an explicit binarization of attention and infinitely
small patches, the Essential Matrix Module can, at best, compute an approximation. We now discuss how close this approximation can get.

The closeness of these approximation depends in part on the network architecture and field of view. Throughout, we use patches
that are arrayed in a $24\times24$ grid.

\begin{figure}
\includegraphics[width=\linewidth]{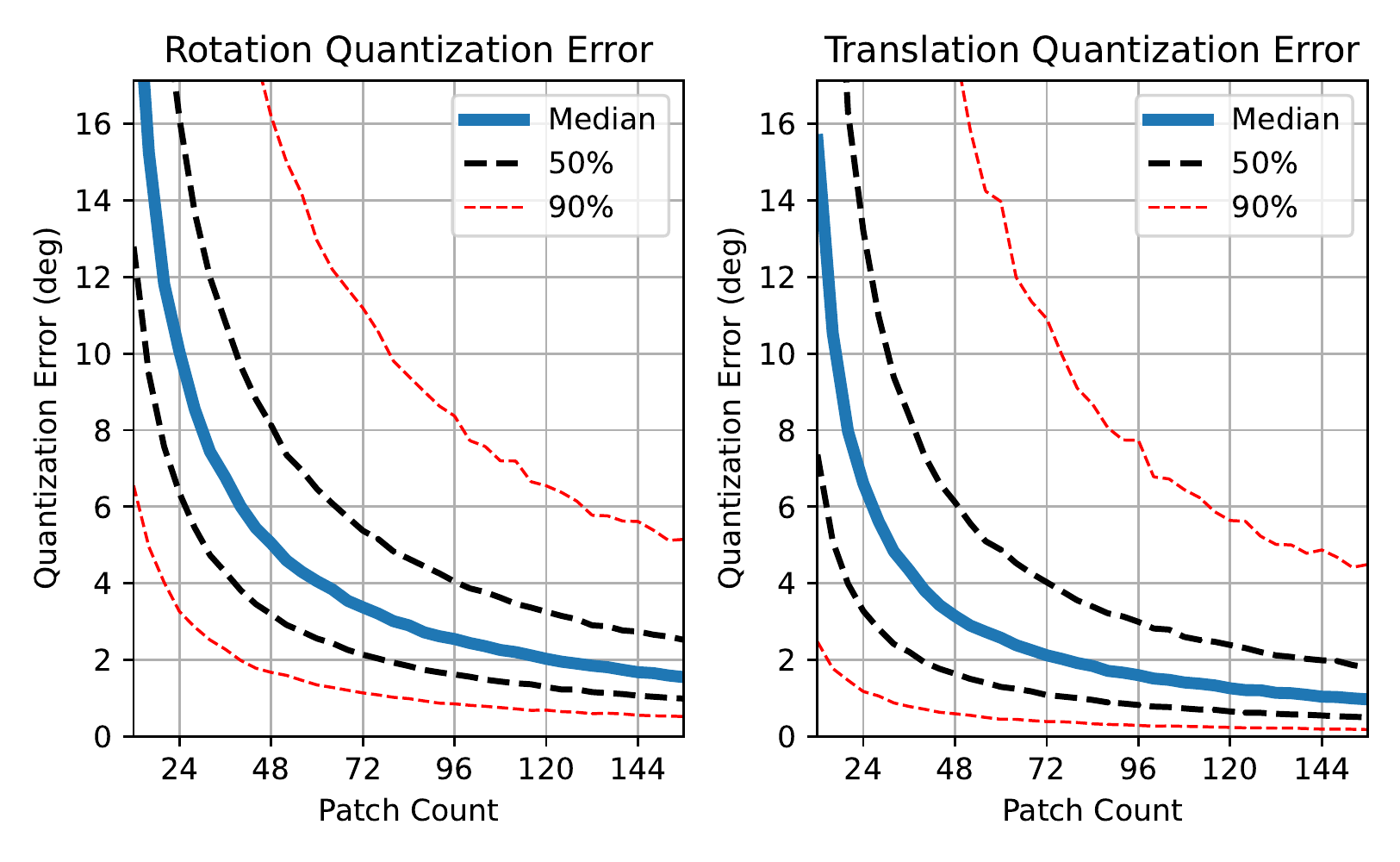}
\caption{Correspondence quantization error as a function of the number of patches for rotation and translation. We use
a patch count of $24\times24$, which has a moderate quantization error. However, since transformers can contain sub-patch
information implicitly, this quantization may be considerably lower.}
\label{fig:quant}
\end{figure}

\parnobf{Representing Each Correspondence as a Patch} 
We replace each correspondence with its equivalent patch, effectively quantizing the correspondence locations. With
patches that are the size of pixels, this has close to no impact on the accuracy of estimating pose; if one represents
the image with a handful of patches, this clearly ought to have a large impact on the accuracy. We now analyze the impact
empirically.

\parnoit{Defining Quantization Error}
For a given quantization level $q$ (i.e., number of patches that uniformly divide the image along each axis),
we generated 10,000 instances of synthetic correspondence by: sampling a
relative camera pose with uniform Euler angles, and translation ${\sim}\textrm{Unif}(-1,1)$, as well as a set of 3D points ${\sim}\textrm{Unif}(-1,1)$. We render these points
to the images using the Matterport3D intrinsics producing a set of correspondences
$(x_i, y_i) \leftrightarrow (x_i', y_i')$. We then compute the relative pose
two ways: first, we do this with the original correspondences, yielding $\RB_o$ and $\tB_o$; 
second, we do it with the correspondences uniformly quantized to $q$ levels, which yields estimates $\RB_q$ and $\tB_q$. We then define the quantization error
as the rotation geodesic between $\RB_o$ and $\RB_q$ as well as the angle between $\tB_o$ and $\tB_q$.

We then plot the median quantization error per quantization level, along with 50\% and 90\% intervals in Figure~\ref{fig:quant}. 
The quantization error rapidly decreases as patch size increases. We use a patch count of $24\times24$ in this work, 
which corresponds to a moderate quantization error ($\textrm{d}(\RB_o, \RB_q) {\approx}10^\circ$, $\textrm{d}(\tB_o, \tB_q) {\approx}7^\circ$).
Transformer tokens can represent sub-patch information, and once the patch count reaches $96\times96$, 
the errors become quite small ($\textrm{d}(\RB_o, \RB_q) {\approx} 2.5^\circ$, $\textrm{d}(\tB_o, \tB_q) {\approx}  1.6^\circ$).

\begin{figure}
\includegraphics[width=\linewidth]{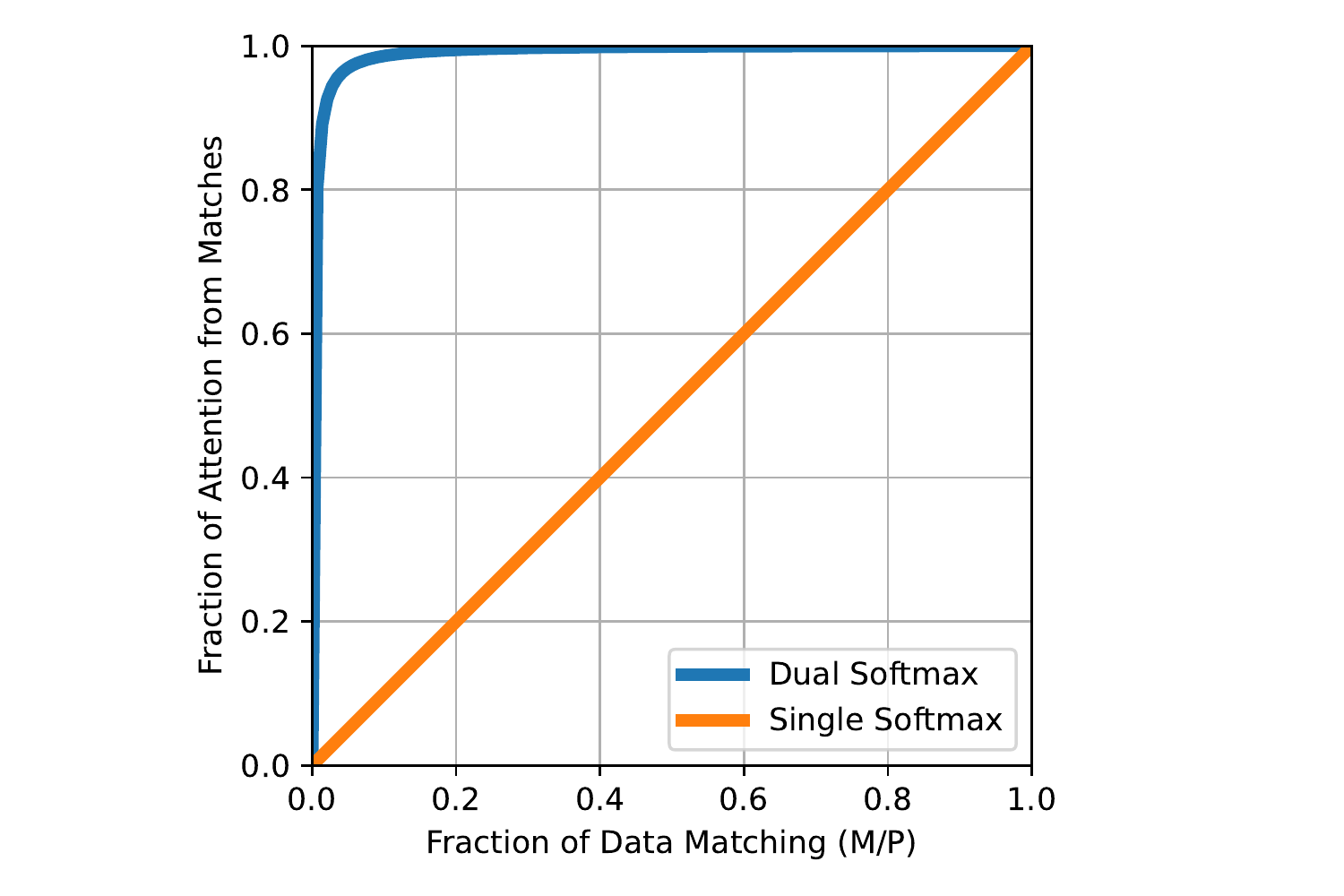}
\caption{Fraction of attention contributed by matches as a fraction of prevalence of matches. Dual softmax enables matches to rapidly dominate the attention matrix's entries.}
\label{fig:fracattention}
\end{figure}

\parnobf{Producing $\AB$} We now discuss how closely a transformer can its attention $\AB = \textrm{norm}(\QB \KB^\top)$ match the binarized matrix
that our setup uses. We divide this into two cases: patches that have correspondence and patches without correspondence.
We refer to the total contribution as the total size of the weights for a patch $j$, or $\sum_{k=1}^P \AB_{j,k}$.

\parnoit{Patches with correspondence} If patch $j$ has a correspondence with
patch $k$, then we would like $\AB_{j,k} = 1$ and $\AB_{j,k'} = 0$ for all $k' \not = k$. 
Standard attention cannot exactly reach this, but can get arbitrarily close by making its dot product
$(\QB \KB^\top)_{\j,k}$ as high as possible. Thus the total contribution of a matching patch $j$ 
$\sum_{k=1}^P \AB_{j,k} {\approx} 1$.

\parnoit{Patches without correspondence} If patch $j$ has no correspondence, then we would like
$\AB_{j,k'} = 0$ for all $k'$. This is impossible under standard attention. We can get this 
to be as close to $0$ as possible, by the following: make 
$(\QB \KB^\top)_{j,:}$ and $(\QB \KB^\top)_{:,j}$ all equal, which in turn
makes the resulting softmax distributions uniform. In turn this makes
$\textrm{softmax}(\QB \KB^\top)_{j,k'} = \textrm{softmax}(\QB \KB^\top)_{k',j} = 1/P$.
Under dual-softmax, then $\AB_{j,k'} = 1/P^2$ for all $k'$.
Thus, the total contribution of a patch $j$ is $\sum_{k'=1}^P 1/P^2 = 1/P$.

Together, this means that with dual softmax, the vast majority of the attention matrix's
energy comes from matches. We do a simple experiment, assuming that
$(\QB \KB^\top)_{j,k} = 100$ if patches $j$ and $k$ match and $1$ otherwise. We can quantify the fraction
of the attention that is from matches by examining
the fraction of the resulting matrix $\AB = \textrm{norm}(\QB \KB^\top)$ that corresponds
to matches. We plot this as a function of the prevalence of matches in Fig.~\ref{fig:fracattention},
comparing regular and dual softmax. With dual softmax, a handful of matches dominate the attention,
whereas for regular softmax, attention increases linearly.

%% file: supp_synth.tex
\parnobf{Datasets}
For each dataset, we generate an instance consisting of a scene and relative camera pose. These can be used to derive
features that are suitable for training. {\it Scenes:} Each scene consists of a points drawn uniformly inside a sphere with center $\cB$ with its
coordinates independently and identically sampled from $\textrm{Unif}(-\frac{1}{2},\frac{1}{2})$ and radius 
$r {\sim}\textrm{Unif}(\frac{1}{2},\frac{3}{2})$. This sampling is done with rejection sampling.
{\it Relative Camera Pose:} We generate Euler angles ($\theta_x, \theta_y, \theta_z$) for the three axes, and a translation vector $\tB$.
In all cases, we reject samples with $||\tB|| \le \frac{1}{2}$.
\begin{enumerate}
\item {\bf 3D} (General 3D Motion): $\theta_x, \theta_y, \theta_z \sim \textrm{Unif}(0,360^\circ)$; $\tB_x, \tB_y, \tB_z \sim \textrm{Unif}(-1,1)$.
\item {\bf 2D Large} (Large-Rotation Motion In XZ plane): $\theta_y \sim \textrm{Normal}(0,25^\circ)$, $\theta_x, \theta_z \sim \textrm{Normal}(0,1.25^\circ)$;
$\tB_x, \tB_z \sim \textrm{Normal}(0,\frac{1}{3})$, $\tB_y \sim \textrm{Normal}(0,\frac{1}{60})$.
\item {\bf 2D Medium} (Medium-Rotation Motion In XZ plane): $\theta_y \sim \textrm{Normal}(0,5^\circ)$, $\theta_x, \theta_z \sim \textrm{Normal}(0,0.25^\circ)$;
$\tB_x, \tB_z \sim \textrm{Normal}(0,\frac{1}{3})$, $\tB_y \sim \textrm{Normal}(0,\frac{1}{60})$.
\item {\bf 2D Small} (Small-Rotation Motion In XZ plane): $\theta_y \sim \textrm{Normal}(0,1^\circ)$, $\theta_x, \theta_z \sim \textrm{Normal}(0,0.05^\circ)$;
$\tB_x, \tB_z \sim \textrm{Normal}(0,\frac{1}{3})$, $\tB_y \sim \textrm{Normal}(0,\frac{1}{60})$.
\end{enumerate}

Given a scene and relative camera pose, we project the points onto a virtual camera with height and width 800 units, a focal length of 800 units and principal point of 400 
units. We record the point if it is in front of the camera, and on the virtual sensor. We reject the image pair and scene if fewer than 100 points out of 10K random points 
are valid for both cameras.

\parnobf{Input Feature}
Given a 3D point, we denote its projection into image 1 as $\xB$ and its projection into image 2 as $\xB'$. Given the set of valid points, we compute the
explicit form of $\UB^\top \UB$ with two modifications. First, for numerical stability
we divide each coordinate by the width of the image and subtract $1/2$, which centers the data. Second, 
to make the feature independent of the number of correspondences, we normalize by the number of points to obtain $\frac{1}{N} \UB^\top \UB$ rather than $\UB^\top \UB$. These
are identical from the perspective of eigenvectors, but normalizing makes the feature indepdendent of the number of points.

\parnobf{Method}
For each task, we train a multilayer perceptron consisting of 3 hidden layers with 4096 units each. Each hidden layer is capped with a leaky ReLU. We predict a 
normalized vector (3D for translation direction, 4D for rotation).

%% file: supp_matrix.tex
To enable visually verifying Equation~\ref{eqn:uniqueentires}, 
we'll show a visual expansion. 
To avoid notational clutter, we will drop the $i$th index
and deal with $\xB = [u,v,1]$ and $\xB' = [u',v',1]$.
Our goal is to show that $(\xB \otimes \xB')^\top (\xB \otimes \xB')$
has the same entries as $\phi(\xB)\phi(\xB')^\top$.
We'll first expand out $\phi(\xB) \phi(\xB')^\top$. When factored
out,
\begin{equation}
\label{eqn:eight}
\phi(\xB) \phi(\xB')^\top = \left[
\begin{array}{ccccccccc}
\textcolor[rgb]{0.75,0.0,0.0}{1} & \textcolor[rgb]{0.75,0.0,0.0}{{u'}} & \textcolor[rgb]{0.75,0.0,0.0}{{v'}} & \textcolor[rgb]{0.75,0.0,0.0}{{u'}{v'}} & \textcolor[rgb]{0.75,0.0,0.0}{{{u'}^2}} & \textcolor[rgb]{0.75,0.0,0.0}{{{v'}^2}}\\ 
\textcolor[rgb]{0.0,0.75,0.0}{{u}} & \textcolor[rgb]{0.0,0.75,0.0}{{u}{u'}} & \textcolor[rgb]{0.0,0.75,0.0}{{u}{v'}} & \textcolor[rgb]{0.0,0.75,0.0}{{u}{u'}{v'}} & \textcolor[rgb]{0.0,0.75,0.0}{{u}{{u'}^2}} & \textcolor[rgb]{0.0,0.75,0.0}{{u}{{v'}^2}}\\ 
\textcolor[rgb]{0.0,0.0,0.75}{{v}} & \textcolor[rgb]{0.0,0.0,0.75}{{v}{u'}} & \textcolor[rgb]{0.0,0.0,0.75}{{v}{v'}} & \textcolor[rgb]{0.0,0.0,0.75}{{v}{u'}{v'}} & \textcolor[rgb]{0.0,0.0,0.75}{{v}{{u'}^2}} & \textcolor[rgb]{0.0,0.0,0.75}{{v}{{v'}^2}}\\ 
\textcolor[rgb]{0.75,0.75,0.0}{{u}{v}} & \textcolor[rgb]{0.75,0.75,0.0}{{u}{v}{u'}} & \textcolor[rgb]{0.75,0.75,0.0}{{u}{v}{v'}} & \textcolor[rgb]{0.75,0.75,0.0}{{u}{v}{u'}{v'}} & \textcolor[rgb]{0.75,0.75,0.0}{{u}{v}{{u'}^2}} & \textcolor[rgb]{0.75,0.75,0.0}{{u}{v}{{v'}^2}}\\ 
\textcolor[rgb]{0.0,0.75,0.75}{{u^2}} & \textcolor[rgb]{0.0,0.75,0.75}{{u^2}{u'}} & \textcolor[rgb]{0.0,0.75,0.75}{{u^2}{v'}} & \textcolor[rgb]{0.0,0.75,0.75}{{u^2}{u'}{v'}} & \textcolor[rgb]{0.0,0.75,0.75}{{u^2}{{u'}^2}} & \textcolor[rgb]{0.0,0.75,0.75}{{u^2}{{v'}^2}}\\ 
\textcolor[rgb]{0.75,0.0,0.75}{{v^2}} & \textcolor[rgb]{0.75,0.0,0.75}{{v^2}{u'}} & \textcolor[rgb]{0.75,0.0,0.75}{{v^2}{v'}} & \textcolor[rgb]{0.75,0.0,0.75}{{v^2}{u'}{v'}} & \textcolor[rgb]{0.75,0.0,0.75}{{v^2}{{u'}^2}} & \textcolor[rgb]{0.75,0.0,0.75}{{v^2}{{v'}^2}}\\ 
\end{array}
\right].
\end{equation}
We color the terms according to which row they appear in $\phi(\xB) \phi(\xB')^\top$.
The matrix created for each correspondence is $(\xB \otimes {\xB'})^\top (\xB \otimes {\xB'})$. We'll first define
$(\xB \otimes {\xB'})$:
\begin{equation}
(\xB \otimes {\xB}') = \left[ \begin{array}{ccccccccc} \textcolor[rgb]{0.0,0.75,0.0}{{u}{u'}} & \textcolor[rgb]{0.0,0.75,0.0}{{u}{v'}} & \textcolor[rgb]{0.0,0.75,0.0}{{u}} & \textcolor[rgb]{0.0,0.0,0.75}{{v}{u'}} & \textcolor[rgb]{0.0,0.0,0.75}{{v}{v'}} & \textcolor[rgb]{0.0,0.0,0.75}{{v}} & \textcolor[rgb]{0.75,0.0,0.0}{{u'}} & \textcolor[rgb]{0.75,0.0,0.0}{{v'}} & \textcolor[rgb]{0.75,0.0,0.0}{1}\end{array} \right].\end{equation}
We can then compute the outer product 
$(\xB \otimes {\xB'})^\top (\xB \otimes {\xB'})$. This is highly redundant -- note that the
ith row and ith column are identical. More specifically, 
\begin{equation}
\label{eqn:ten}
(\xB \otimes {\xB}')^\top (\xB \otimes {\xB}')= \left[
\begin{array}{ccccccccc}
\textcolor[rgb]{0.0,0.75,0.75}{{u^2}{{u'}^2}} & \textcolor[rgb]{0.0,0.75,0.75}{{u^2}{u'}{v'}} & \textcolor[rgb]{0.0,0.75,0.75}{{u^2}{u'}} & \textcolor[rgb]{0.75,0.75,0.0}{{u}{v}{{u'}^2}} & \textcolor[rgb]{0.75,0.75,0.0}{{u}{v}{u'}{v'}} & \textcolor[rgb]{0.75,0.75,0.0}{{u}{v}{u'}} & \textcolor[rgb]{0.0,0.75,0.0}{{u}{{u'}^2}} & \textcolor[rgb]{0.0,0.75,0.0}{{u}{u'}{v'}} & \textcolor[rgb]{0.0,0.75,0.0}{{u}{u'}}\\ 
\textcolor[rgb]{0.0,0.75,0.75}{{u^2}{u'}{v'}} & \textcolor[rgb]{0.0,0.75,0.75}{{u^2}{{v'}^2}} & \textcolor[rgb]{0.0,0.75,0.75}{{u^2}{v'}} & \textcolor[rgb]{0.75,0.75,0.0}{{u}{v}{u'}{v'}} & \textcolor[rgb]{0.75,0.75,0.0}{{u}{v}{{v'}^2}} & \textcolor[rgb]{0.75,0.75,0.0}{{u}{v}{v'}} & \textcolor[rgb]{0.0,0.75,0.0}{{u}{u'}{v'}} & \textcolor[rgb]{0.0,0.75,0.0}{{u}{{v'}^2}} & \textcolor[rgb]{0.0,0.75,0.0}{{u}{v'}}\\ 
\textcolor[rgb]{0.0,0.75,0.75}{{u^2}{u'}} & \textcolor[rgb]{0.0,0.75,0.75}{{u^2}{v'}} & \textcolor[rgb]{0.0,0.75,0.75}{{u^2}} & \textcolor[rgb]{0.75,0.75,0.0}{{u}{v}{u'}} & \textcolor[rgb]{0.75,0.75,0.0}{{u}{v}{v'}} & \textcolor[rgb]{0.75,0.75,0.0}{{u}{v}} & \textcolor[rgb]{0.0,0.75,0.0}{{u}{u'}} & \textcolor[rgb]{0.0,0.75,0.0}{{u}{v'}} & \textcolor[rgb]{0.0,0.75,0.0}{{u}}\\ 
\textcolor[rgb]{0.75,0.75,0.0}{{u}{v}{{u'}^2}} & \textcolor[rgb]{0.75,0.75,0.0}{{u}{v}{u'}{v'}} & \textcolor[rgb]{0.75,0.75,0.0}{{u}{v}{u'}} & \textcolor[rgb]{0.75,0.0,0.75}{{v^2}{{u'}^2}} & \textcolor[rgb]{0.75,0.0,0.75}{{v^2}{u'}{v'}} & \textcolor[rgb]{0.75,0.0,0.75}{{v^2}{u'}} & \textcolor[rgb]{0.0,0.0,0.75}{{v}{{u'}^2}} & \textcolor[rgb]{0.0,0.0,0.75}{{v}{u'}{v'}} & \textcolor[rgb]{0.0,0.0,0.75}{{v}{u'}}\\ 
\textcolor[rgb]{0.75,0.75,0.0}{{u}{v}{u'}{v'}} & \textcolor[rgb]{0.75,0.75,0.0}{{u}{v}{{v'}^2}} & \textcolor[rgb]{0.75,0.75,0.0}{{u}{v}{v'}} & \textcolor[rgb]{0.75,0.0,0.75}{{v^2}{u'}{v'}} & \textcolor[rgb]{0.75,0.0,0.75}{{v^2}{{v'}^2}} & \textcolor[rgb]{0.75,0.0,0.75}{{v^2}{v'}} & \textcolor[rgb]{0.0,0.0,0.75}{{v}{u'}{v'}} & \textcolor[rgb]{0.0,0.0,0.75}{{v}{{v'}^2}} & \textcolor[rgb]{0.0,0.0,0.75}{{v}{v'}}\\ 
\textcolor[rgb]{0.75,0.75,0.0}{{u}{v}{u'}} & \textcolor[rgb]{0.75,0.75,0.0}{{u}{v}{v'}} & \textcolor[rgb]{0.75,0.75,0.0}{{u}{v}} & \textcolor[rgb]{0.75,0.0,0.75}{{v^2}{u'}} & \textcolor[rgb]{0.75,0.0,0.75}{{v^2}{v'}} & \textcolor[rgb]{0.75,0.0,0.75}{{v^2}} & \textcolor[rgb]{0.0,0.0,0.75}{{v}{u'}} & \textcolor[rgb]{0.0,0.0,0.75}{{v}{v'}} & \textcolor[rgb]{0.0,0.0,0.75}{{v}}\\ 
\textcolor[rgb]{0.0,0.75,0.0}{{u}{{u'}^2}} & \textcolor[rgb]{0.0,0.75,0.0}{{u}{u'}{v'}} & \textcolor[rgb]{0.0,0.75,0.0}{{u}{u'}} & \textcolor[rgb]{0.0,0.0,0.75}{{v}{{u'}^2}} & \textcolor[rgb]{0.0,0.0,0.75}{{v}{u'}{v'}} & \textcolor[rgb]{0.0,0.0,0.75}{{v}{u'}} & \textcolor[rgb]{0.75,0.0,0.0}{{{u'}^2}} & \textcolor[rgb]{0.75,0.0,0.0}{{u'}{v'}} & \textcolor[rgb]{0.75,0.0,0.0}{{u'}}\\ 
\textcolor[rgb]{0.0,0.75,0.0}{{u}{u'}{v'}} & \textcolor[rgb]{0.0,0.75,0.0}{{u}{{v'}^2}} & \textcolor[rgb]{0.0,0.75,0.0}{{u}{v'}} & \textcolor[rgb]{0.0,0.0,0.75}{{v}{u'}{v'}} & \textcolor[rgb]{0.0,0.0,0.75}{{v}{{v'}^2}} & \textcolor[rgb]{0.0,0.0,0.75}{{v}{v'}} & \textcolor[rgb]{0.75,0.0,0.0}{{u'}{v'}} & \textcolor[rgb]{0.75,0.0,0.0}{{{v'}^2}} & \textcolor[rgb]{0.75,0.0,0.0}{{v'}}\\ 
\textcolor[rgb]{0.0,0.75,0.0}{{u}{u'}} & \textcolor[rgb]{0.0,0.75,0.0}{{u}{v'}} & \textcolor[rgb]{0.0,0.75,0.0}{{u}} & \textcolor[rgb]{0.0,0.0,0.75}{{v}{u'}} & \textcolor[rgb]{0.0,0.0,0.75}{{v}{v'}} & \textcolor[rgb]{0.0,0.0,0.75}{{v}} & \textcolor[rgb]{0.75,0.0,0.0}{{u'}} & \textcolor[rgb]{0.75,0.0,0.0}{{v'}} & \textcolor[rgb]{0.75,0.0,0.0}{1}\\ 
\end{array}
\right]
\end{equation}
Note that the rows of $\phi(\xB) \phi(\xB')^\top$ appear in $3 \times 3$ blocks inside 
$(\xB \otimes {\xB'})^\top (\xB \otimes {\xB'})$. In particular, inside each $3 \times 3$ block, the 
columns of the $\phi(\xB) \phi(\xB')^\top$ appear in the following order:
\begin{equation}
\left[ \begin{array}{ccc}
5 & 4 & 2 \\4 & 6 & 3 \\2 & 3 & 1 \\\end{array} \right].
\end{equation}